\documentclass[11pt, letterpaper, logo]{appier-ai-research}

\PassOptionsToPackage{comma,numbers,sort,compress}{natbib}
\usepackage[comma,numbers,sort,compress]{natbib}
\hypersetup{colorlinks=true, citecolor=magenta}

\usepackage{enumitem}
\usepackage{url}
\usepackage{graphicx}
\usepackage{wrapfig}
\usepackage{capt-of}
\usepackage{booktabs}
\usepackage{titletoc}
\usepackage{bm}

\pdftrailerid{redacted}

\title{TasteRoute: Personalized Routing for Video Generation}

\correspondingauthor{ray.eed08g@nctu.edu.tw, cylin@appier.com}

\author[$\dagger$, 1]{Zhi Rui Tam}
\author[2]{Chao-Chung Wu}
\author[2]{Sin-Han Yang}
\author[2]{Peyton Ku}
\author[2]{Brendan Kuang}
\author[2]{Tzu-Ting Hsieh}
\author[2]{Min-Fang Hsu}
\author[2]{Fang-Ling Tsai}
\author[1]{Yun-Nung Chen}
\author[3]{Wei-Chiu Ma}
\author[$\dagger$, 2]{Chieh-Yen Lin}

\affil[1]{National Taiwan University}
\affil[2]{Appier Inc.}
\affil[3]{Cornell University}
\affil[$\dagger$]{Corresponding authors}

\begin{abstract}
Rapid progress in video generation has led to a plethora of models that differ substantially in capability and generation cost. This raises a natural question: can each request be efficiently routed to an appropriate model?
We find that even when the consensus of the other annotators is used as an oracle, it agrees with each annotator's own favorite only 34--55\% of the time.
Motivated by this observation, we introduce TasteRoute, a personalized video-generation router that selects a generator jointly based on the input request, user preferences, and available generation budget. Across text-to-video and image-to-video settings, TasteRoute is competitive with strong simple baselines on preference routing while reducing average generation cost. The cost saving increases under higher budget caps. Finally, we release TasteRoute-3k, a human-annotated dataset containing multi-model video comparisons, quality judgments, preference rankings, and user-profile signals to facilitate future research on personalized and cost-aware video routing.
\end{abstract}

\begin{document}

\maketitle

\section{Introduction}
Recent advances in conditional video generation have led to models with increasingly high visual quality and controllability \citep{hacohen2026ltx, gao2025seedance, team2025kling, googledeepmind2025veo}. With just a few clicks, users can now generate videos with remarkable visual fidelity. Nevertheless, in practice, models capable of producing comparably high-quality videos can differ substantially in generation cost. Their strengths may also vary: some may produce finer visual detail, while others may better preserve physical plausibility or achieve a desired style. 
This raises a natural question: which model should a user choose for a given request?

At first glance, this question may seem ill-posed, as model selection involves many intricate, entangled factors. The choice becomes more manageable, however, when viewed from the user's perspective: does a model produce an output that meets their needs, and which qualities matter most to them?
Indeed, if the goal is to generate a short transition shot, clear content and coherent motion may have higher priority than  fine-grained visual details. Users may therefore seek an output that meets their needs rather than maximize quality along every dimension~\citep{schwartz2002maximizing}.
Additionally, individual preferences shape the value of different outputs: one user may favor a stylized rendering, while another prefers photorealism. Even when two models meet the same basic requirements, their outputs may appeal to different users. 
Model selection should therefore account for both task-specific requirements and individual preferences to determine whether a more expensive model delivers value that justifies its cost.

In this paper, we study the problem of personalized video generation routing. 
Given a generation request in the form of text and an optional reference image, our goal is to determine which video model to use for a particular user under a given budget. 
The router makes this choice before generation, considering both whether a model is likely to produce an acceptable output for the input request and how well that output may align with the user's preferences. %, represented through a learned user embedding.

The task is far from trivial, in part because learning such a router requires data that captures both sides of this decision. To learn user preferences, we need judgments that preserve individual choices rather than aggregate them into a single consensus ranking. To learn which models suit a given request, we need comparisons of outputs from different generators under the same input. These comparisons must also distinguish basic acceptability from subjective preference, so that clear generation failures do not obscure differences in taste among otherwise usable outputs. 
Unfortunately, such data remains scarce due to heavy reliance on human annotations. 

To this end, we carefully collect \textbf{TasteRoute-3K},\footnote{Available at \url{https://huggingface.co/datasets/appier-ai-research/tasteroute-3k}.} a human-annotated dataset with professional written prompts and videos generated by SOTA models that combines quality assessments, diverse preference rankings, and user profiles for videos generated by multiple models under the same request.
We find that annotators agree more often when video comparisons involve unacceptable artifacts, but differ substantially when choosing among acceptable ones. Moreover, the preferred generator varies with both the request and the annotator, confirming that routing must account for both. 
Using this dataset, we further develop \textbf{TasteRoute}, a personalized router that combines representations of the input text and optional image with a learned user embedding. The router predicts basic acceptability alongside a preference score for each candidate generator, allowing it to distinguish shared quality requirements from individual preferences. At inference time, these predictions guide the selection of a generator within the available budget.

Across text-to-video and image-to-video settings, we evaluate TasteRoute against generalized and personalized routing. In generalized routing, TasteRoute achieves performance comparable to a strong kNN baseline. In the personalization setting, TasteRoute-Personal reduces the average generation cost by 8--12\% at budget caps of \$1.0--\$2.0 while achieving higher routing accuracy than any other personalized baseline.
We further study questionnaire-conditioned routing as a cold-start variant.
However, personalization offers limited benefits under tight budgets and remains challenging when users express directly conflicting preferences.

\section{Related Work}
\subsection{Model Routing}
The idea of model-level routing, which selects a model for inference based on the input, dates back to classification settings \citep{kilic2011learning, aljundi2017expert}, where routing was primarily used to maximize predictive performance. With the emergence of generative models such as LLMs and diffusion models, inference cost has become an additional consideration because serving increasingly capable models requires substantial computation. To maximize quality under a limited cost budget, \citet{chen2024frugalgpt, ICLR2024_b47d93c9} study cascade-based routing, while \citet{hu2024routerbench} formalizes standardized quality--cost evaluation for model routers. Subsequent work generalizes routing through richer representations and decision mechanisms, including query-based representations \citep{chen2024routerdc, li2026rethinkingpredictivemodelingllm}, graph-based query--model interaction \citep{feng2025graphrouter}, multi-armed bandits \citep{rezaei2025more}, sequential multi-round routing \citep{zhang2025routerr}, and training-free routing \citep{tang2026vdar}. Beyond optimizing quality and cost, model routing can also support personalization by selecting different models for different users \citep{dai2025personalizedrouter, Piskala_2024}.

\subsection{Model Routing for Generative Models}

Most work on model routing targets language models, but some early work has begun to formulate model selection directly for image generation. PAK-UCB \citep{pmlr-v267-hu25m} casts prompt-aware generator selection as an online contextual-bandit problem under sequential feedback, using automatic fidelity metrics such as CLIPScore as rewards. More recently, OctoT2I \citep{jiang2026octot2iselfevolvingagentictexttoimage} moves beyond one-shot selection with an agentic, self-evolving framework that autonomously discovers capability profiles across T2I models and uses verifier feedback to guide subsequent routing decisions.

Compared with language and image generation, model-level routing for video generation remains unexplored. This setting is particularly suited to routing because video generators differ substantially in both capability and serving cost, with reported 1080p image-to-video prices spanning roughly \$0.30--\$24.00 per generated minute \citep{artificialanalysis2026video}, while different users may also prefer different generators for the same request. These disparities motivate routing policies that select video generators according to the request and user rather than relying on a single backend.

\subsection{Video Preference Datasets}
Existing video-preference datasets are not well suited for training a video router. They either
evaluate earlier models \citep{liu2026improving, he2025videoscore2} whose outputs are often dominated by
obvious artifacts, or contain relatively few human preference annotations \citep{jiang2026vga}. For example,
\citet{huang2026vajudger} contains only 1,150 human-annotated pairs, with the remainder labeled by a
calibrated Gemini~3.1 judge. Crowd-sourced leaderboards such as Artificial Analysis's Video Arena
collect feedback across a broader range of generators \citep{artificialanalysis2026video} but do not release
their data.

\section{Problem Formulation}
\label{sec:problem}
We study video-generation routing, where a router selects a generator for each text-to-video (T2V) or image-to-video (I2V) request under a generation budget.
Given a request and a pool of candidate generators with known costs, the router scores every candidate and selects the highest-scoring one whose cost is within the budget. In personalized routing, the score additionally depends on the user.

Learning such a router requires multi-generator video comparisons under the same request, preferences recorded at the individual level, and a separate assessment of video acceptability.
Therefore we construct \textbf{TasteRoute-3k}, a human-annotated video-routing dataset built around these requirements. 

\section{TasteRoute-3K: Data Collection and Annotation}
\label{gen_inst}
\label{sec:data}
We begin by collecting 277 video-generation prompts, used as text-only inputs for text-to-video (T2V) and paired with a reference image for image-to-video (I2V). Approximately half of the prompts are drawn from an existing collection \citep{he2025videoscore2}, while the rest are written by professional video creators with more demanding requirements, such as timestamp-specific motion and camera movement. A prompt may be used in both settings, giving 394 requests (274 T2V and 120 I2V). We then generate candidate videos from a catalog of 30 generator endpoints, covering 16:9 and 9:16 aspect ratios with durations of 4, 6, and 10 seconds, for a total of 3{,}128 annotated videos. Table~\ref{tab:dataset_stats} summarizes the dataset.

\begin{table}[tb]
\caption{TasteRoute-3K at a glance. A \emph{pool} is one request in one setting together
with every generator labeled for it; a \emph{group} is the unit one annotator ranks at a
time. A \emph{comparison} is one ordered pair of videos within a group; the consensus
outcome of a pair is the label chosen by the most annotators, and pairs with no strict
plurality are \emph{ambiguous} and excluded from the consensus targets.}
\label{tab:dataset_stats}
\begin{center}
\begin{small}
\begin{tabular}{lrl}
\toprule
\bf Quantity & \bf Count & \bf Definition \\
\midrule
Distinct prompts            & 277   & unique prompt texts \\
Pools                       & 394   & one per request, i.e.\ prompt $\times$ setting (274 T2V, 120 I2V); 393 evaluable \\
Groups                      & 812   & 2.06 per pool, at most six videos each \\
Annotated videos            & 3{,}128 & distinct (pool, generator) generations \\
Generator endpoints         & 30    & catalog of Appendix~\ref{app:models}; 28 win at least one pool \\
Annotators                  & 9     & four general annotators, five professional video creators \\
\midrule
Consensus pairs             & 8{,}181 & distinct video pairs with an aggregated outcome \\
\quad non-ambiguous         & 7{,}062 & training targets after dropping ambiguous pairs \\
Doubly-judged comparisons   & 19{,}338 & (comparison, annotator pair) with two decisive votes \\
\bottomrule
\end{tabular}
\end{small}
\end{center}
\end{table}

\subsection{Annotation Pipeline}

Nine annotators (four general, five professional video creators) participate in the labeling. Before labeling, all annotators must first complete a calibration meeting on the interface and rubric, covering details such as what a text artifact looks like. We deliberately do not calibrate the aesthetic preference since it is the signal that we wish to capture for personalization.

Three units recur throughout the paper. The \emph{catalog} is the full list of 30 generator endpoints (Appendix~\ref{app:models}); the \emph{pool} of a request is the generators whose outputs are labeled for it; and a \emph{group} is what one annotator ranks at a time, at most six videos drawn from one pool. A large pool is therefore covered by several overlapping groups; the group is the unit annotators work in, and the pool is the unit of every aggregation and evaluation that follows.

Our annotation pipeline separates video quality acceptability from subjective preference.  % Annotators first assess whether each generation is technically acceptable based on several criteria, then rank acceptable candidates according to their own preferences. In addition, we collect user preference profiles to support personalized routing. (comment this out and let the following paragraphs do the work)
Annotators first evaluate each video's quality using a shared rubric, then rank candidate videos according to their individual preferences. 

\paragraph{Video Quality and Acceptability} In the first stage, annotators assess each video along four rubric criteria: (i)~\textbf{Text integrity}: whether visible text is malformed, hallucinated, or unreadable; (ii)~\textbf{Motion consistency}: whether motion shows smearing, discontinuities, or implausible movement; (iii)~\textbf{Subject realism}: do subjects have misformed features or physically implausible rendering; and (iv)~\textbf{Instructions following}: does the video follow the conditioning prompt.

Each criterion is rated \emph{reject}, \emph{borderline}, or \emph{accept}, as is overall acceptability, giving five quality fields per video. The acceptability score remains per-annotator and is never pooled such that a video one annotator rejects may still be ranked by another (Section~\ref{sec:analysis} analyzes these disagreements). 

\paragraph{Human Subjective Preference} 

Following quality assessment, annotators rank candidate videos according to their individual preferences, with ties allowed. 
Each group includes two anchor videos shared with the other groups of the same pool, which connect their rankings.
We prioritize comparisons between similarly ranked generators to better resolve their relative preferences, while randomly sampling a subset of groups to maintain broader coverage of generator pairs.

\paragraph{User Preference Profiles} 
Each annotator also completes a five-item questionnaire covering self-reported preferences for video qualities (e.g., prompt accuracy, natural motion, etc).
Because self-reported preferences may not reflect actual choices, each annotator also completes four reference-video comparisons, choosing between videos A and B, indicating no preference, or selecting cannot judge. 
After selecting either A or B, annotators additionally report their preference strength and up to two perceptual factors (e.g. motion naturalness, camera framing, etc.). These responses provide the cold-start preference signal for video routing (Section~\ref{sec:conditioning}). 

\subsection{General Preference Aggregation}
\label{sec:preference_aggregation}

For the general routing task we aggregate annotations within each pool. 
Each cross-generator video pair takes the outcome that the most annotators chose among win, loss and tie, 
and carries the number of annotators behind it as its weight; 
a pair with no strict plurality is ambiguous and is dropped. 
On the rest we fit a weighted Bradley–Terry (BT) model, 
one per connected component of each pool's comparison graph, 
with ties counted as half a win and weights normalized by the number of comparisons the originating listwise ranking induces. 
Its win probabilities are the soft ranking targets of Section~\ref{sec:objectives}, yielding 7,062 consensus training pairs. 
We train on these probabilities because they place every generator of a pool on one smooth scale, 
but score every analysis and evaluation with the consensus utility of Section~\ref{sec:analysis}, 
which needs no model fit and keeps the ground truth independent of the training target.

\paragraph{Availability.}
Each row of TasteRoute-3K is one annotated pool: the prompt, the reference image for I2V,
the generated videos, and the ranking collected for it. The dataset is available at
\begin{center}
\vspace{-0.5\baselineskip}
\url{https://huggingface.co/datasets/appier-ai-research/tasteroute-3k}
\vspace{-0.5\baselineskip}
\end{center}

\section{When Does Video Routing Need Personalization?}
\label{sec:analysis}
With this dataset in hand, we examine what a video router needs to learn: (i) how generator strengths depend on the request, (ii) which preferences are shared across annotators, and (iii) when individual preferences diverge.
Within a pool, each video pair takes the outcome that the most annotators chose (Section~\ref{sec:preference_aggregation}), and each generator receives a \emph{consensus utility}: its win rate over the pairs it appears in, counting a win as $1$ and a tie as $0.5$, with each pair weighted by the number of annotators behind it. The top-scoring generator is the \emph{consensus winner}.

\begin{figure}[t]
  \centering
  \begin{minipage}[b]{0.506\textwidth}
    \centering
    \includegraphics[width=\linewidth]{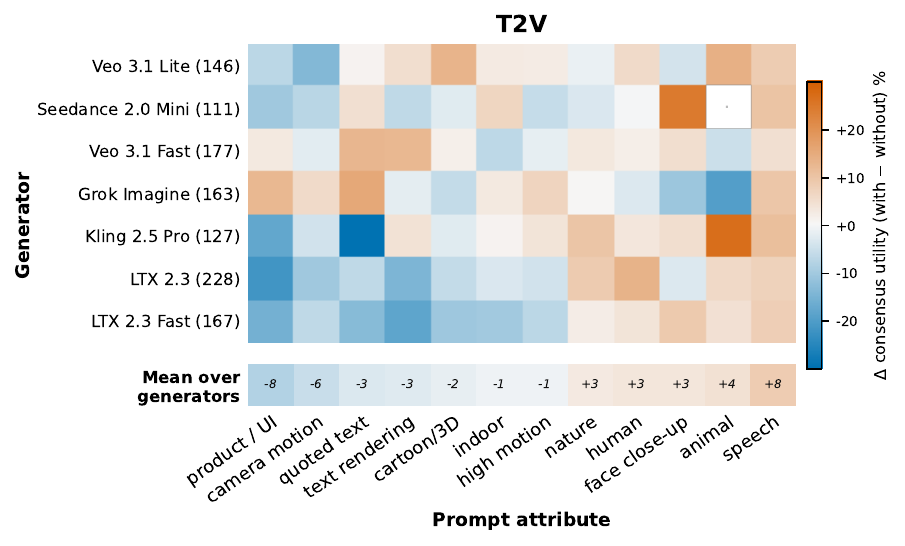}
  \end{minipage}\hfill
  \begin{minipage}[b]{0.474\textwidth}
    \centering
    
      \includegraphics[width=\linewidth]{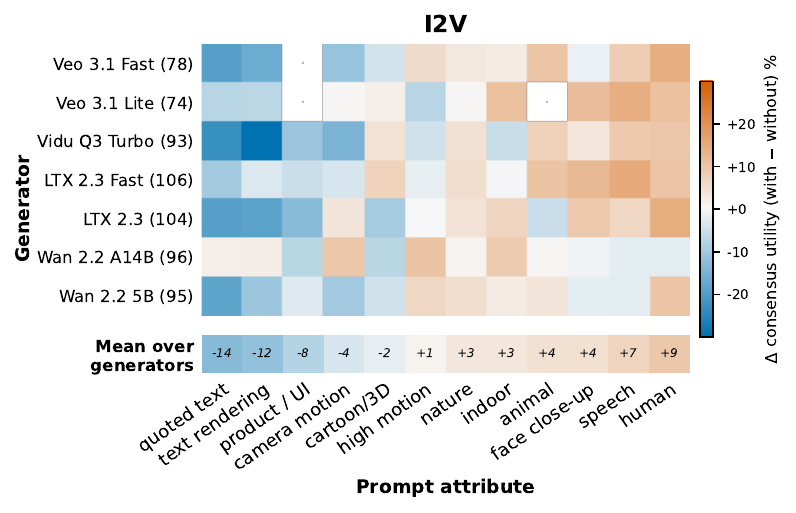}
    
  \end{minipage}
  
  \caption{\textbf{The best generator changes with the prompt, in both settings.}
  Each cell is how much better or worse a generator scores on prompts with a given
  attribute than on prompts without it (\%); the bottom strip is the mean over
  generators. \textbf{Left (T2V)} and \textbf{right (I2V)} use the same twelve
  attributes and the same estimator. Different generators lead on different attributes,
  and the leader changes on 5 of the 12 attributes in T2V and 8 of the 12 in I2V.}
  
  \label{fig:attribute_generator}
  
\end{figure}

\paragraph{No generator wins everywhere, and which one wins depends on the prompt.}
If one generator consistently performed best, a fixed choice would suffice. 
We therefore examine how generator performance varies across prompt attributes (e.g.\ animation style)
in Figure~\ref{fig:attribute_generator} (left):
each cell shows the difference between a generator's mean consensus utility on prompts with a given attribute and without the given attribute.
In T2V, Veo~3.1~Lite has the highest consensus utility overall, yet it loses 5 of the 12 attributes to three rivals: Seedance~2.0~Mini on face close-ups, Kling~2.5~Pro on animals, 
and Grok~Imagine on product / UI, quoted text and camera motion.
A near-zero mean score means the prompt changes
\emph{who} wins, which is what a router can exploit.

 \providecommand{\disagreecaption}{}\renewcommand{\disagreecaption}{\textbf{The consensus winner is nobody's favorite.} One T2V
pool in which no video was rejected; the first column is the pool's consensus rank, the rest
are the individual annotators'.}

\begin{figure}[t]
  \centering
  \includegraphics[width=0.46\textwidth]{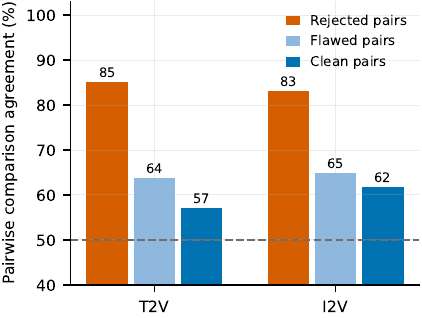}
  \caption{\textbf{Annotators agree on bad videos but barely on good ones.} How often two
  annotators who judged the same pair chose the same video, for pairs where one video was
  rejected, where one carried a defect, and where both were clean.}
  \label{fig:agreement_quality}
\end{figure}

\begin{figure}[tb]
  \centering
  \includegraphics[width=0.60\textwidth]{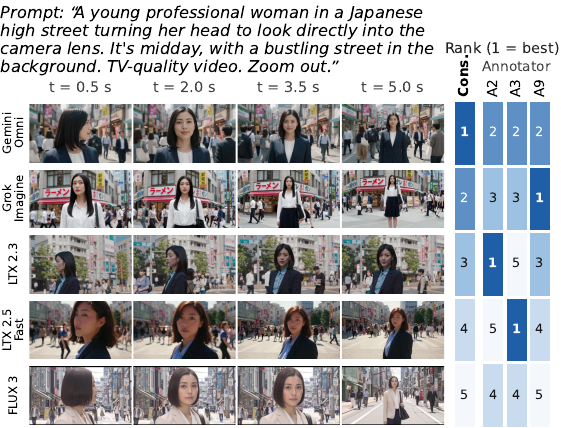}
  \caption{\disagreecaption}
  \label{fig:disagreement_consensus}
\end{figure}

\paragraph{Annotators agree on what is unacceptable but not on which is the best.}
\emph{Agreement} is how often two annotators who judged the same video pair prefer the same
video. We measure it on 19{,}338 \emph{doubly-judged comparisons} over 376 pools and nine
annotators, each one a video pair with two annotators who both expressed a decisive preference.  Figure~\ref{fig:agreement_quality}{} splits these pairs
by annotator-assessed acceptability and defect flags. 
When either annotator rejected either video, annotators agree 83--85\% of the time: bad output is bad for everyone. Agreement then decreases as the videos get better, to 64--66\% when both are acceptable but one carries a defect flag, and to 57\% (T2V) and 62\% (I2V) when both are clean, barely above the 50\% of a coin flip. 
Quality explains agreement, but prompt content does not: of 16 video attributes, only 
cartoon or 3D prompts robustly raise agreement on clean pairs ($+18\%$). Prompts that 
ask for exact on-screen text show the largest negative trend ($-10\%$). 
See Figure~\ref{fig:agreement_attributes} in Appendix~\ref{app:analysis} for the full
breakdown.

\paragraph{Even when all videos are acceptable, the consensus ranking may not match any annotator’s top choice.}
Figure~\ref{fig:disagreement_consensus} showcases a pool in which every video passed every annotator's acceptability check, yet the consensus winner is nobody's first choice: all three rank it second. A harder pool has three favorites in a cycle (Grok $\succ$ FLUX $\succ$ LTX~2.5~Fast $\succ$ Grok), so no aggregation rule can name a winner that more than one annotator endorses (Figure~\ref{fig:disagreement_cycle}, Appendix~\ref{app:analysis}). 
Across pools, 58\% have a unique consensus winner, and 7\% contain a majority-preference cycle.

\begin{figure}[tb]
  \centering
  \includegraphics[width=0.75\textwidth]{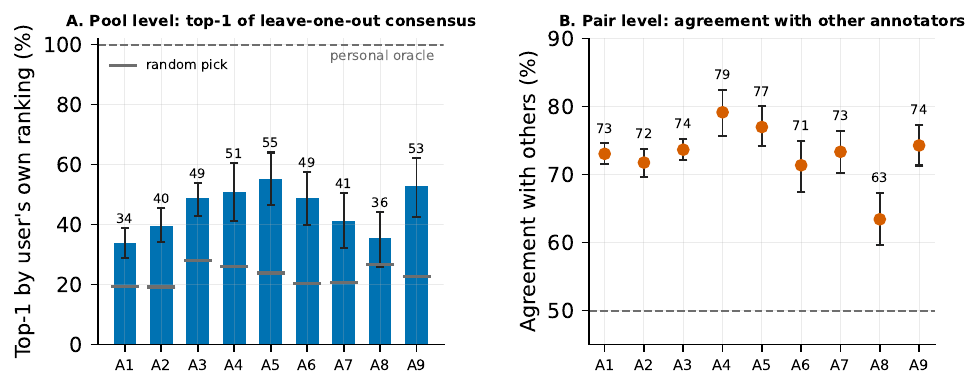}
  \caption{\textbf{How far consensus gets each annotator.} (A)~Pool level: how often the
  favorite of the \emph{other} annotators on the same pool is also this annotator's own
  top choice (tie-aware). Gray marks a random pick; the dashed line at 100\%
  (\emph{personal oracle}) is a router that always matches the annotator. (B)~Pair
  level: how often the annotator prefers the same video as another annotator in a
  single comparison. Bars in both panels are pool-level bootstrap 95\% intervals.}
  \label{fig:headroom}
\end{figure}

\paragraph{The pool's favorite is the individual's favorite only about half the time.}
For every pool an annotator labeled,
we take the option preferred by the \emph{other} annotators and check whether it matches
this annotator’s own top choice; no model training is involved. The consensus favorite
matches the annotator’s own choice 34--55\% of the time, against 19--28\% for a random
pick. 
The consensus is a much better guess than random and still a poor guess for any particular annotator, which is the gap user conditioning has to close.
The mismatch is observed across all nine annotators: 
every annotator agrees with the rest on 63--79\% of these comparisons (Figure~\ref{fig:headroom}).

These findings motivate a routing framework that captures both shared and individual preferences. 
A generalized router learns how generator preferences vary across prompts from pooled 
judgments, while a personalized router additionally models individual preferences among 
acceptable outputs. A single architecture supports both by conditioning on the prompt and, 
when available, user information (Section~\ref{sec:conditioning}). The BT soft 
targets of Section~\ref{sec:preference_aggregation} represent pooled preferences 
probabilistically rather than reducing them to hard winner--loser labels.

\section{TasteRoute: Personalized Video Routing}
\label{router}
Section~\ref{sec:analysis} split preferences into a shared part (what is
acceptable) and a personal part (which acceptable output a user prefers).
TasteRoute scores every candidate generator from one frozen embedding of the
request, without generating a video, and one architecture yields both a
generalized router and per-user routers trained on a user's own feedback. We
describe the architecture and how it conditions on a user
(Section~\ref{sec:architecture}), then the objectives of the two variants
(Section~\ref{sec:objectives}).
\subsection{Architecture}
\label{sec:architecture}
Each video generator request is first encoded using the multimodal \texttt{Qwen3-VL-Embedding-2B} encoder~\citep{qwen3vlembedding}, taking text prompt for T2V and both text prompt and reference image for I2V. Video generator is represented by a learned embedding.
\label{sec:conditioning}
For a user seen in training, we learn $e_u$ along with a shared general-user embedding ${\bm{g}}$, which stands in whenever no user information is available (general routing).
To prevent overfitting on single individual users, we apply user dropout with probability $p_{\mathrm{drop}}=0.4$ of replacing user embedding with ${\bm{g}}$.
For a new user, we instead encode the onboarding questionnaire as text with the same frozen encoder and project it into the user space with a small MLP; this conditioner needs no user-ID table and falls back to ${\bm{g}}$ when no questionnaire exists.

All input features are normalized and concatenated, together with their pairwise products, which are then fed to three small MLP heads: a \emph{general} ranking head that sees only the request and generator, a \emph{personalized} ranking head that additionally sees the user vector, and a shared \emph{ordinal quality} head that predicts the four rubric criteria and acceptability of Section~\ref{sec:data} as ratings in $\{1,2,3\}$. 
The two ranking heads write scores $s^{g}$ and $s^{p}(\cdot,e_u)$, and the quality head writes one latent $q^{(f)}$ per rubric field; throughout, a superscript $g$ marks a quantity computed without user conditioning and $p$ marks one computed with user conditioning.
Routing in Section~\ref{sec:res} ranks candidates by the ranking heads alone; the quality head is evaluated separately.

\subsection{Training Objectives}
\label{sec:objectives}
Every TasteRoute model uses the same two kinds of losses, a pairwise ranking loss and an ordinal
quality loss weighted by $\lambda$. The variants differ in what supervises them: which pairs they see, whose
preference the target comes from, and which ranking branch is active. 
Appendix~\ref{app:impl} lists the remaining differences, including the context features.

\paragraph{Losses.}
For candidates $A,B$ of the same request with scores $s_A,s_B$ and a target preference $p_{AB}$,
\begin{equation}
    \mathcal{L}_{\mathrm{rank}} = -w\!\left[
        p_{AB}\log\sigma\!\left(\tfrac{s_A - s_B}{\tau}\right)
        + (1 - p_{AB})\log\sigma\!\left(\tfrac{s_B - s_A}{\tau}\right)
    \right],
    \label{eq:ranking_loss}
\end{equation}
with temperature $\tau$ and the pair weight $w$ of Section~\ref{sec:preference_aggregation}. 
For the generalized router $p_{AB}$ is the BT win probability aggregated across annotators; for a personalized router it is the annotator's own outcome ($1$, $0$, or $0.5$ for a tie). 
The quality head uses a cumulative-link ordinal loss $\mathcal{L}_{\mathrm{qual}}$: for each annotated rubric field $f$ and threshold $k\in\{1,2\}$, a binary cross-entropy on $P(Y^{(f)}>k)=\sigma(q^{(f)}-\theta^{(f)}_k)$, where $Y^{(f)}$ is the annotator's rating of field $f$ and the $\theta^{(f)}_k$ are learned ordered thresholds, averaged over the fields present.

\paragraph{TasteRoute-Generic.}
One model per data fold, trained on the aggregated pairs
$\mathcal{D}_{\mathrm{BT}}$ and the rubric ratings $\mathcal{Q}$ of all
annotators through the general head only,
\begin{equation}
    \mathcal{L}_{\mathrm{gen}} =
    \mathcal{L}_{\mathrm{rank}}\bigl(s^{g};\mathcal{D}_{\mathrm{BT}}\bigr)
    + \lambda\,\mathcal{L}^{g}_{\mathrm{qual}}(\mathcal{Q}).
    \label{eq:general_loss}
\end{equation}

\paragraph{TasteRoute-Personal.}
One independent model per annotator $u$, trained only on that annotator's
pairs $\mathcal{D}_u$ and ratings $\mathcal{Q}_u$. The same pairs train the
general head without user information and the personalized head with the
(dropout-masked) embedding $\tilde e_u$,
\begin{equation}
    \mathcal{L}_{u} =
    \mathcal{L}_{\mathrm{rank}}\bigl(s^{g};\mathcal{D}_u\bigr)
    + \mathcal{L}_{\mathrm{rank}}\bigl(s^{p}(\cdot,\tilde e_u);\mathcal{D}_u\bigr)
    + \lambda\bigl[\mathcal{L}^{g}_{\mathrm{qual}}(\mathcal{Q}_u)
    + \mathcal{L}^{p}_{\mathrm{qual}}(\mathcal{Q}_u)\bigr].
    \label{eq:personal_loss}
\end{equation}
At inference it scores with $s^{p}(\cdot,e_u)$. No annotator's model sees
another annotator's labels, so Section~\ref{sec:res} measures what one user's
own feedback buys. Two TasteRoute baselines reuse these objectives:
\emph{TasteRoute-Generic-Pooled} trains the general head on the union of all
annotators' individual outcomes without aggregation, and \emph{TasteRoute-Personal-Static}
replaces $e_u$ with the questionnaire embedding. All models use
$\lambda=0.25$, selected on the validation split (Appendix~\ref{app:impl}).

\section{Evaluation Setup}
\label{sec:exp}
\paragraph{Evaluation metrics.}
For the general video routing, we use five-fold cross-validation with prompt-disjoint splits such that no identical prompt appears in both training and validation or test split. Predictions are evaluated against the consensus utilities mentioned in Section~\ref{sec:analysis}. For the personalized video routing, we split each annotator labeling into 15\% validation, 15\% test with the rest as train. We report the average of three random seed runs.

To study label efficiency, each annotator's training prompt groups are ordered by a diversity criterion seeded with the questionnaire embedding, which serves as the first training unit, and each new group is the one farthest from the centroid of the units chosen so far, and a \emph{cap} $c$ keeps only the first $c$ groups; the matched pooled model at cap $c$ receives $c$ groups in total, balanced across annotators. In both settings, a router chooses among the generators of the pool being evaluated, and held-out labels are never used for routing.

\paragraph{Metrics and routing under budget.}
Utility is the per-pool consensus utility defined in Section~\ref{sec:analysis}, computed from individual preferences for personalized evaluation and aggregated across annotators for general evaluation.
We report tie-aware Top-1 (the selection is a co-winner), Top-3 (the selection ranks within the top three), Normalized Discounted Cumulative Gain (NDCG). %with gain $2^u-1$, and the catalog cost of the selected generator.
Under budget cap $B$, routers may select only generators priced at or below $B$, but evaluation retains the full pool as ground truth.
In particular, NDCG@$N_{\mathrm{aff}}$ is normalized against the top $N_{\mathrm{aff}}$ generators in the full pool, where $N_{\mathrm{aff}}$ is the number of affordable generators.
Thus, an unaffordable best generator remains the Top-1 target.
We sweep $B\in[\$0.5,\$2.0]$ for both evaluation settings, using a finer grid up to \$1.5 for personalized evaluation.

\subsection{Baselines}\label{sec:baselines}

\begin{itemize}[leftmargin=1.2em,itemsep=1pt,topsep=2pt]
  \item \textbf{kNN}~\citep{li2026rethinkingpredictivemodelingllm}: retrieves the $k$ most similar training prompts using the same embedding models as TasteRoute and ranks generators by the mean observed utility ($k$ chosen on validation);
  \emph{Same-user kNN} implements the personalized version of kNN routing by limiting to only a single annotator rather than the aggregated mean observed utility in kNN.
  \item \textbf{Octo}~\citep{jiang2026octot2iselfevolvingagentictexttoimage}: an agentic video-adapted image router. It selects the cheapest generator whose remembered pass rate for the prompt's capability bucket exceeds a threshold, then verifies the output, and escalates on rejection for up to four attempts. We evaluate VideoScore2 \citep{he2025videoscore2}, GPT-5.6, Gemini~3.7 Flash as verifiers, and include human labels as signal for upper-bound  (\emph{Octo-Human-Label}).
  \item \textbf{FrugalGPT}~\citep{chen2024frugalgpt}: A video-routing variant of FrugalGPT routes each input through a cascade of video models, first generating a video and then evaluating its quality with a video-quality scorer; if the output is deemed sufficiently reliable, the process stops, otherwise the input is escalated to a more capable, higher-cost model.  
  \item \textbf{Hierarchical User-BT}: first fits a shared generator ranking from all pairwise votes, then it learns a small per-annotator correction score based on each annotator's history preference. The adjustment weight $\lambda$ chosen based on validation set.
  \item \textbf{PersonalizedRouter-GNN}~\citep{dai2025personalizedrouter}: We adopted   graph-based personalized router with video generators as nodes, inputs as query nodes, and user-conditioned link prediction.
  \item \textbf{TasteRoute variants} (Section~\ref{sec:objectives}): 
  \emph{TasteRoute-Generic-Pooled} and \emph{TasteRoute-Personal-Static} for cold start scenario where no user routing preference was recorded.
  \item \textbf{Random}: uniformly selects one generator from the valid candidate set for each input. 
\end{itemize}

The details of implementation can be found in Appendix~\ref{app:impl}.

\section{Performance of Generalized and Personalized Routing}
\label{sec:res}
As Section~\ref{sec:analysis} separates video routing into consensus and personalized layers, in this section, we train consensus preferences alone for generalized (user-agnostic) routing, and then include annotator preferences to improve personalized routing. We also study the needed data amount per-annotator to trade off labeling effort against routing quality .

\subsection{Generalized Routing Results}
\begin{figure*}[t]
  \centering
  \includegraphics[width=0.96\textwidth]{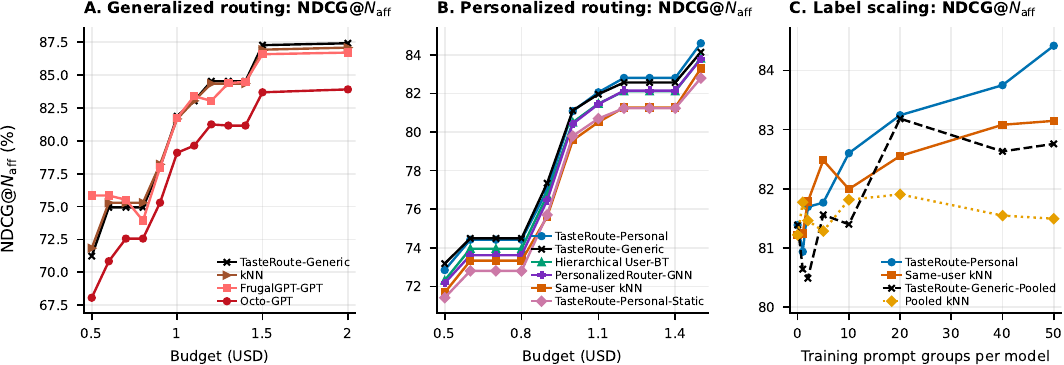}
  \caption{\textbf{Routing quality by limiting the cost available to the router in two settings}: generalized routing (A) and personalized routing (B). In both settings, TasteRoute (Generic, Personal) performs competitively against the strong kNN baseline.
  (C) We also compare training dataset scaling across all routers and find that TasteRoute-Personal scales better as data continues to grow. All performance is measured using NDCG.}
  \label{fig:routing-ndcg}
\end{figure*}

Figure~\ref{fig:routing-ndcg} reports NDCG for this subsection and the next: panel~A grades generalized routing, panels~B and~C personalized routing. Top-1 and Top-3 give the same picture and are in Appendix~\ref{app:metrics}.

\paragraph{Learned routing offers modest gains over retrieval.} Figure~\ref{fig:routing-ndcg}A shows that TasteRoute-Generic and kNN achieve similar NDCG across budget caps. 
At the \$2.0 cap, TasteRoute-Generic reaches 42.0\% Top-1, compared with 39.7\% for kNN, 38.2\% for FrugalGPT, and 33.6\% for Octo. 
These results suggest that learning prompt--generator interactions offers limited improvement over retrieval under pooled preferences.
With an oracle verifier using held-out human labels, Octo reaches 58.0\% Top-1, highlighting the role of verification quality in generate-then-verify routing.

\begin{figure}[tb]
  \centering
  \includegraphics[width=0.63\textwidth]{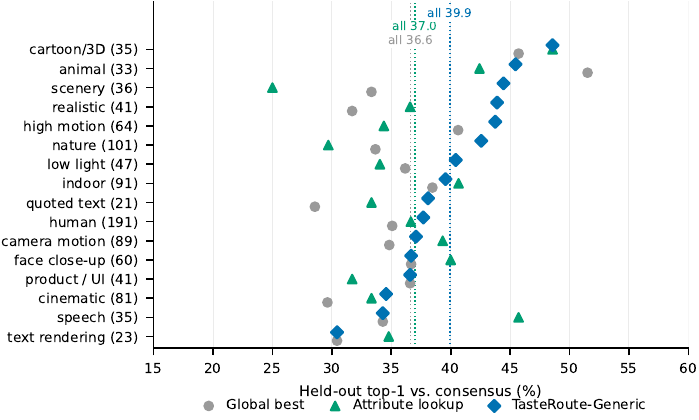}
  \caption{\textbf{TasteRoute-Generic by prompt attribute.} Held-out outer-fold T2V
  pools grouped by the keyword-detected attributes of Section~\ref{sec:analysis}, counts in
  parentheses, all graded against the pool's consensus ranking. The router is compared with
  two prompt-independent references: the single best generator on the training folds
  (\emph{global best}) and the best generator for each attribute on the training folds
  (\emph{attribute lookup}). Markers are offset vertically within a row and each dotted
  line is that series' rate over all attributes. Note that
  Figure~\ref{fig:case_study_personal} uses a different ground truth (each annotator's own
  ranking) and a different unit (annotator--pool pairs), so the two are not comparable.}
  \label{fig:case_study_generalized}
\end{figure}

\paragraph{Prompt attributes alone are insufficient for generator selection.}
Next we examine whether a simpler attribute-based lookup can replace a learned router. Figure~\ref{fig:case_study_generalized} shows that generator preferences vary across prompt attributes, with TasteRoute-Generic's Top-1 ranging from 48.6\% on cartoon and 3D prompts to 30.4\% on prompts requesting rendered text. 
Prompt attributes provide some useful routing signal, yet a per-attribute lookup table achieves only 37.0\% Top-1 accuracy, compared with 36.6\% for a single global-best generator and 39.9\% for TasteRoute-Generic.
This suggests that the learned router exploits prompt-dependent information beyond coarse attribute categories. 
To examine its sensitivity to request, we modify one attribute phrase while keeping the remaining prompt unchanged. 
These edits alter TasteRoute-Generic's top-ranked generator in 5--55\% of cases across ten attributes, with an average flip rate of 25\%.

\subsection{Personalized Routing Results}

\paragraph{Personalization becomes more useful as the budget expands.}
Figure~\ref{fig:routing-ndcg}B shows that both TasteRoute variants perform competitively with kNN, Hierarchical User-BT, and PersonalizedRouter-GNN across budget caps. TasteRoute-Personal performs similarly to TasteRoute-Generic below \$1.1 per video but pulls ahead as the budget increases. At lower caps, the limited choice of affordable generators leaves less room for individual preferences to influence routing. For example, an annotator's preferred generator is unaffordable in approximately 20\% of pools at \$0.5, compared with almost none at \$1.5. As more generators become affordable, TasteRoute-Personal can better exploit differences between individual and pooled preferences. 

\paragraph{How much user feedback is needed for personalization?} To study personalization with limited user feedback, we vary the number of labeled prompt groups used for training (Figure~\ref{fig:routing-ndcg}C). At each training size $c$, TasteRoute-Personal uses $c$ groups from one annotator, while TasteRoute-Generic-Pooled uses the same number of groups across annotators. With fewer than 10 groups, the Generalized router and Same-user kNN perform better. TasteRoute-Personal surpasses the pooled router at approximately 10 groups and Same-user kNN by 50. Its NDCG continues to improve at the largest training size, suggesting that additional user feedback may further improve personalization.

\begin{figure}[t]
  \centering
  \begin{minipage}[t]{0.505\textwidth}
    \centering
    \includegraphics[width=\linewidth]{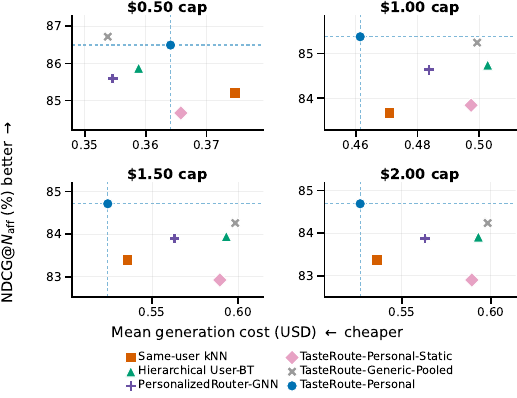}
    \caption{\textbf{Cost against routing quality.} Upper left is cheaper and better;
    dashed lines mark the personalized router. Quality here is NDCG over the affordable
    pool, not over the full pool of Figure~\ref{fig:routing-ndcg}.}
    \label{fig:cost_quality}
  \end{minipage}\hfill
  \begin{minipage}[t]{0.475\textwidth}
    \centering
    \includegraphics[width=\linewidth]{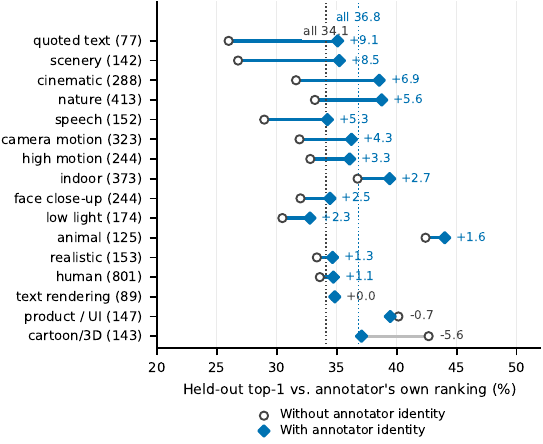}
    \caption{\textbf{Personalization by prompt attribute.} One router scored with and
    without the annotator's identity, graded against that annotator's own ranking. The bar
    is the gain; dotted lines are the two overall rates.}
    \label{fig:case_study_personal}
  \end{minipage}
\end{figure}

\section{Analysis on Personalized Router}
Building on the overall performance gains, we further analyze how personalized routing is improved from generic routing in terms of generation cost and alignment with individual preferences.

\subsection{Trade-offs between Cost and Preference}

\paragraph{The preferred generators are also the cheaper ones.}

Figure~\ref{fig:cost_quality} plots each router's mean cost against routing quality at every budget cap. NDCG of routing quality is calculated over the affordable pool instead of the full pool used in Figure~\ref{fig:routing-ndcg}. At the \$1.0, \$1.5, and \$2.0 budget caps, TasteRoute-Personal is the cheapest and most accurate router, reducing cost by 8\% to 12.0\% compared to the most expensive alternatives at the same caps. Only at the \$0.5 budget cap, 
TasteRoute-Generic-Pooled is marginally ahead on both axes. 
TasteRoute is trained without cost supervision, and the budget cap is only an affordability filter at inference time, so this saving is a byproduct of the learned preference ranking: the preferred generators are not the expensive ones.

\paragraph{Learning on cost harms accuracy.}
Saving cost is also learnable: following ~\citet{li2026rethinkingpredictivemodelingllm}, we applied a user-defined trade-off parameter governing the preference for performance versus cost, the full sweep over parameters and budget caps is given in Table~\ref{tab:personalized_cost_tradeoff}.
A parameter small enough makes Top-1 unchanged at every cap and saves at most 0.03\%. Larger parameters trade accuracy for cost: at the \$1.5 cap a parameter value of $0.3$ reduces cost by 9.8\% for 1.4 points of Top-1, and $1.0$ by 25.2\% for 2.8 points. This gives us an operating point to trade-off between explicit accuracies and generation costs.

These runs are exploratory: they use the fixed
T2V/I2V subset of the personalization test splits and a single seed triple, so they
indicate the shape of the trade-off rather than a tuned operating point.

\begin{table}[tb]
\centering
\small
\setlength{\tabcolsep}{4pt}
\caption{Cost-preference trade-off for the personalized TasteRoute.
Top-1 accuracy(\%) and cost savings(\%) are reported, averaged over nine annotators and three split seeds on a fixed T2V/I2V subset of the personalization test splits. Savings are relative to unchanged training ($\lambda=0$) at the same budget cap. The utility-score formula is unchanged; $\lambda$ weights only the additional training cost loss.}
\label{tab:personalized_cost_tradeoff}
\begin{tabular}{c*{4}{rr}}
\toprule
 & \multicolumn{2}{c}{\$0.50 cap}
 & \multicolumn{2}{c}{\$1.00 cap}
 & \multicolumn{2}{c}{\$1.50 cap}
 & \multicolumn{2}{c}{\$1.70 cap} \\
\cmidrule(lr){2-3}\cmidrule(lr){4-5}
\cmidrule(lr){6-7}\cmidrule(lr){8-9}
$\lambda$
 & Top-1\% $\uparrow$ & Saving \%$\uparrow$
 & Top-1\% $\uparrow$ & Saving \%$\uparrow$
 & Top-1\% $\uparrow$ & Saving \%$\uparrow$
 & Top-1\% $\uparrow$ & Saving \%$\uparrow$ \\
\midrule
0     & 47.01 & 0.00 & 42.51 & 0.00  & 39.85 & 0.00  & 39.85 & 0.00  \\
\midrule
0.003 & 47.01 & 0.00 & 42.51 & 0.03  & 39.85 & 0.03  & 39.85 & 0.03  \\
0.01  & 47.01 & 0.00 & 42.51 & 0.03  & 39.85 & 0.03  & 39.85 & 0.03  \\
0.03  & 46.92 & 0.13 & 42.33 & 0.84  & 39.69 & 0.73  & 39.69 & 0.73  \\
0.1   & 46.25 & 0.91 & 40.95 & 2.86  & 38.79 & 3.82  & 38.79 & 3.81  \\
0.3   & 46.09 & 2.31 & 40.26 & 7.33  & 38.44 & 9.77  & 38.44 & 9.73  \\
0.5   & 45.48 & 4.18 & 39.27 & 9.52  & 37.34 & 16.21 & 37.34 & 16.57 \\
1.0   & 45.61 & 9.34 & 38.92 & 16.66 & 37.05 & 25.18 & 37.05 & 25.50 \\
\bottomrule
\end{tabular}
\end{table}

\subsection{Understanding Personalization Gains}
\label{sec:learned}
\begin{figure}[tb]
  \centering
  \includegraphics[width=0.50\textwidth]{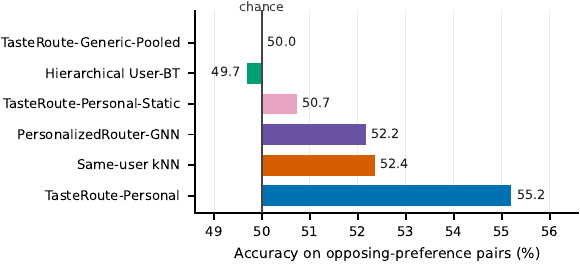}
  \caption{\textbf{Recovering opposing preferences.} Accuracy on held-out video pairs where
  two annotators disagree. A router without user information scores 50\% by construction.}
  \label{fig:preference_recovery}
\end{figure}

We first test whether personalization captures annotator-specific preferences using held-out video pairs on which two annotators express opposite preferences. TasteRoute-Generic makes the same prediction for both annotators and achieves 50\% accuracy by construction. TasteRoute-Personal reaches 55.2\%, compared with 52.4\% for Same-user kNN and 52.2\% for PersonalizedRouter-GNN, indicating modest sensitivity to individual preferences (Figure~\ref{fig:preference_recovery}). We then examine whether this sensitivity translates into larger routing gains when annotators disagree.

\paragraph{Personalization gains are larger when annotators share a preferred generator.}
We partition 120 matched annotator--pool cohorts by whether any annotators share a top-ranked generator. Among the 54 cohorts with a shared winner, personalization improves Top-1 from 36.7\% to 50.1\%; among the remaining 66 cohorts, it improves Top-1 from 25.9\% to 28.7\%, with the latter improvement not statistically conclusive. Thus, although personalization distinguishes some conflicting pairwise preferences, its larger routing gains occur where preferred generators overlap.

\paragraph{How do personalization gains vary across prompt attributes?}
In a separate held-out T2V analysis, we compare the general and annotator-conditioned heads of the same jointly trained router against each annotator’s own ranking (Figure~\ref{fig:case_study_personal}). Personalization raises overall Top-1 from 34.1\% to 36.8\%. Gains range from $-5.6$ percentage points for cartoon/3D and $-0.7$ for product/UI to $+8.5$ for scenery and $+9.1$ for quoted text. The largest gains occur in the two groups with the lowest general-head accuracy, highlighting personalization’s value where general routing is weakest.

\section{Conclusion}
\label{sec:conclusion}
We presented TasteRoute-3k, a human-annotated dataset of multi-generator video comparisons, and TasteRoute, a personalized video-generation router that selects a generator from the request, the user's preferences and the available budget. TasteRoute-3k shows that preferences over video generators are both diverse and conflicting across annotators, and that a router conditioned on annotator identity improves over generalized routing once the budget no longer constrains the choice. TasteRoute outperforms every personalized baseline at every budget cap and, from the \$1.0 cap upward, is simultaneously the cheapest and the most accurate router. Our analyses of opposite-preference pairs, of the cost--preference trade-off, and of the link between prompt attributes and personalization show that the router encodes annotator-specific preference but mostly corrects the consensus rather than resolving conflicts between users. Closing that gap, with user-conditioned features that raise accuracy without raising generation cost, is the direction we see for future work, alongside understanding how human preference over generated video differs from preference over text and images.
\section{Limitation}
\label{sec:limitation}
The gap between the best router and the oracle remains large (42.0\% against 58.0\% Top-1 at the \$2.0 cap), and widens as more generators and larger budgets make preferences harder to predict. Beyond this, our personalized router mostly corrects the consensus rather than resolving conflicts between annotators (55.2\% on opposite-preference pairs), results rest on nine annotators and a fixed generator catalog, and the observed cost saving reflects these annotators' preference for cheaper generators rather than a general property.

\subsection*{Acknowledgements}
We thank our colleagues at Appier AI Research for feedback on the annotation
interface and on earlier drafts of this paper.

\clearpage
\bibliographystyle{abbrvnat}
\bibliography{references}

\begin{thebibliography}{26}
\providecommand{\natexlab}[1]{#1}
\providecommand{\url}[1]{\texttt{#1}}
\expandafter\ifx\csname urlstyle\endcsname\relax
  \providecommand{\doi}[1]{doi: #1}\else
  \providecommand{\doi}{doi: \begingroup \urlstyle{rm}\Url}\fi

\bibitem[Aljundi et~al.(2017)Aljundi, Chakravarty, and
  Tuytelaars]{aljundi2017expert}
R.~Aljundi, P.~Chakravarty, and T.~Tuytelaars.
\newblock Expert gate: Lifelong learning with a network of experts.
\newblock In \emph{2017 IEEE Conference on Computer Vision and Pattern
  Recognition (CVPR)}, pages 7120--7129. IEEE, 2017.

\bibitem[{Artificial Analysis}(2026)]{artificialanalysis2026video}
{Artificial Analysis}.
\newblock Video generation arena leaderboard.
\newblock \url{https://artificialanalysis.ai/video/arena}, 2026.
\newblock URL \url{https://artificialanalysis.ai/video/arena}.
\newblock Accessed: August 2026.

\bibitem[Chen et~al.(2024{\natexlab{a}})Chen, Zaharia, and
  Zou]{chen2024frugalgpt}
L.~Chen, M.~Zaharia, and J.~Zou.
\newblock Frugal{GPT}: How to use large language models while reducing cost and
  improving performance.
\newblock \emph{Transactions on Machine Learning Research}, 2024{\natexlab{a}}.
\newblock URL \url{https://openreview.net/forum?id=cSimKw5p6R}.

\bibitem[Chen et~al.(2024{\natexlab{b}})Chen, Jiang, Lin, Kwok, and
  Zhang]{chen2024routerdc}
S.~Chen, W.~Jiang, B.~Lin, J.~Kwok, and Y.~Zhang.
\newblock Router{DC}: Query-based router by dual contrastive learning for
  assembling large language models.
\newblock In \emph{The Thirty-eighth Annual Conference on Neural Information
  Processing Systems}, 2024{\natexlab{b}}.
\newblock URL \url{https://openreview.net/forum?id=7RQvjayHrM}.

\bibitem[Dai et~al.(2025)Dai, Feng, and You]{dai2025personalizedrouter}
Z.~Dai, T.~Feng, and J.~You.
\newblock Personalizedrouter: Personalized llm routing via graph-based user
  preference modeling.
\newblock 2025.

\bibitem[Ding et~al.(2024)Ding, Mallick, Wang, Sim, Mukherjee, R\"{u}hle,
  Lakshmanan, and Awadallah]{ICLR2024_b47d93c9}
D.~Ding, A.~Mallick, C.~Wang, R.~Sim, S.~Mukherjee, V.~R\"{u}hle,
  L.~Lakshmanan, and A.~H. Awadallah.
\newblock Hybrid llm: Cost-efficient and quality-aware query routing.
\newblock In \emph{International Conference on Learning Representations}, 2024.
\newblock URL
  \url{https://proceedings.iclr.cc/paper_files/paper/2024/file/b47d93c99fa22ac0b377578af0a1f63a-Paper-Conference.pdf}.

\bibitem[Feng et~al.(2025)Feng, Shen, and You]{feng2025graphrouter}
T.~Feng, Y.~Shen, and J.~You.
\newblock Graphrouter: A graph-based router for {LLM} selections.
\newblock In \emph{The Thirteenth International Conference on Learning
  Representations}, 2025.
\newblock URL \url{https://openreview.net/forum?id=eU39PDsZtT}.

\bibitem[Gao et~al.(2025)Gao, Guo, Hoang, Huang, Jiang, Kong, Li, Li, Li, Li,
  et~al.]{gao2025seedance}
Y.~Gao, H.~Guo, T.~Hoang, W.~Huang, L.~Jiang, F.~Kong, H.~Li, J.~Li, L.~Li,
  X.~Li, et~al.
\newblock Seedance 1.0: Exploring the boundaries of video generation models.
\newblock \emph{arXiv preprint arXiv:2506.09113}, 2025.

\bibitem[{Google DeepMind}(2025)]{googledeepmind2025veo}
{Google DeepMind}.
\newblock {Veo}: A text-to-video generation system.
\newblock Technical report, Google DeepMind, May 2025.
\newblock URL
  \url{https://storage.googleapis.com/deepmind-media/veo/Veo-3-Tech-Report.pdf}.
\newblock Veo 3 Technical Report.

\bibitem[HaCohen et~al.(2026)HaCohen, Brazowski, Chiprut, Bitterman, Kvochko,
  Berkowitz, Shalem, Lifschitz, Moshe, Porat, et~al.]{hacohen2026ltx}
Y.~HaCohen, B.~Brazowski, N.~Chiprut, Y.~Bitterman, A.~Kvochko, A.~Berkowitz,
  D.~Shalem, D.~Lifschitz, D.~Moshe, E.~Porat, et~al.
\newblock Ltx-2: Efficient joint audio-visual foundation model.
\newblock \emph{arXiv preprint arXiv:2601.03233}, 2026.

\bibitem[He et~al.(2026)He, Jiang, Nie, Liu, Jiang, Su, Ma, Lin, Ye, Lu, Wu,
  Schneider, Do, Li, Jia, Zhang, Cheng, Wang, Zhou, Lin, Zhang, Zhang, Huang,
  and Chen]{he2025videoscore2}
X.~He, D.~Jiang, P.~Nie, M.~Liu, Z.~Jiang, M.~Su, W.~Ma, J.~Lin, C.~Ye, Y.~Lu,
  K.~Wu, B.~Schneider, Q.~D. Do, Z.~Li, Y.~Jia, Y.~Zhang, G.~Cheng, H.~Wang,
  W.~Zhou, Q.~Lin, Y.~Zhang, G.~Zhang, W.~Huang, and W.~Chen.
\newblock Videoscore2: Think before you score in generated video evaluation.
\newblock \emph{Transactions on Machine Learning Research}, 2026.
\newblock URL \url{https://openreview.net/forum?id=MpkVh4jH44}.

\bibitem[Hu et~al.(2024)Hu, Bieker, Li, Jiang, Keigwin, Ranganath, Keutzer, and
  Upadhyay]{hu2024routerbench}
Q.~J. Hu, J.~Bieker, X.~Li, N.~Jiang, B.~Keigwin, G.~Ranganath, K.~Keutzer, and
  S.~K. Upadhyay.
\newblock Routerbench: A benchmark for multi-{LLM} routing system.
\newblock In \emph{Agentic Markets Workshop at ICML 2024}, 2024.
\newblock URL \url{https://openreview.net/forum?id=IVXmV8Uxwh}.

\bibitem[Hu et~al.(2025)Hu, Leung, and Farnia]{pmlr-v267-hu25m}
X.~Hu, H.-F. Leung, and F.~Farnia.
\newblock {PAK}-{UCB} contextual bandit: An online learning approach to
  prompt-aware selection of generative models and {LLM}s.
\newblock In \emph{Proceedings of the 42nd International Conference on Machine
  Learning}, 2025.

\bibitem[Huang et~al.(2026)Huang, Tu, Yan, Yang, Han, Hang, Jiang, and
  Wu]{huang2026vajudger}
Y.~Huang, S.~Tu, X.~Yan, Z.~Yang, J.~Han, X.~Hang, Y.-G. Jiang, and Z.~Wu.
\newblock Va-judger: Reward modeling from human preference feedback for joint
  video-audio generation.
\newblock \emph{arXiv preprint arXiv:2608.18607}, 2026.

\bibitem[Jiang et~al.(2026{\natexlab{a}})Jiang, Zheng, Qiao, Huang, Wang, Bo,
  Peng, Chen, Zhou, and Jin]{jiang2026vga}
L.~Jiang, D.~Zheng, Q.~Qiao, H.~Huang, H.~Wang, Y.~Bo, B.~Peng, J.~Chen,
  J.~Zhou, and X.~Jin.
\newblock Vga-bench: A unified benchmark and multi-model framework for video
  aesthetics and generation quality evaluation.
\newblock In \emph{Proceedings of the IEEE/CVF Conference on Computer Vision
  and Pattern Recognition (CVPR)}, 2026{\natexlab{a}}.

\bibitem[Jiang et~al.(2026{\natexlab{b}})Jiang, Chen, Li, Duan, Wang, and
  Zhang]{jiang2026octot2iselfevolvingagentictexttoimage}
X.~Jiang, B.~Chen, G.~Li, Y.~Duan, R.~Wang, and J.~Zhang.
\newblock Octot2i: A self-evolving agentic text-to-image router.
\newblock In \emph{Proceedings of the IEEE/CVF Conference on Computer Vision
  and Pattern Recognition (CVPR)}, 2026{\natexlab{b}}.

\bibitem[Kilic and Alpaydin(2011)]{kilic2011learning}
E.~Kilic and E.~Alpaydin.
\newblock Learning the areas of expertise of classifiers in an ensemble.
\newblock \emph{Procedia Computer Science}, 3:\penalty0 74--82, 2011.

\bibitem[Li et~al.(2026)Li, Zhang, Long, Keqin, Song, Bai, Yang, Xie, Yang,
  Liu, Zhou, and Lin]{qwen3vlembedding}
M.~Li, Y.~Zhang, D.~Long, C.~Keqin, S.~Song, S.~Bai, Z.~Yang, P.~Xie, A.~Yang,
  D.~Liu, J.~Zhou, and J.~Lin.
\newblock Qwen3-vl-embedding and qwen3-vl-reranker: A unified framework for
  state-of-the-art multimodal retrieval and ranking.
\newblock \emph{arXiv preprint arXiv:2601.04720}, 2026.

\bibitem[Li(2026)]{li2026rethinkingpredictivemodelingllm}
Y.~Li.
\newblock Rethinking predictive modeling for llm routing: When simple knn beats
  complex learned routers, 2026.
\newblock URL \url{https://arxiv.org/abs/2505.12601}.

\bibitem[Liu et~al.(2026)Liu, Liu, Liang, Yuan, Liu, Zheng, Wu, Wang, Xia,
  Wang, et~al.]{liu2026improving}
J.~Liu, G.~Liu, J.~Liang, Z.~Yuan, X.~Liu, M.~Zheng, X.~Wu, Q.~Wang, M.~Xia,
  X.~Wang, et~al.
\newblock Improving video generation with human feedback.
\newblock \emph{Advances in Neural Information Processing Systems}, 2026.

\bibitem[Piskala et~al.(2024)Piskala, Raajaa, Mishra, and Bozza]{Piskala_2024}
D.~B. Piskala, V.~Raajaa, S.~Mishra, and B.~Bozza.
\newblock Optiroute dynamic llm routing and selection based on user
  preferences: Balancing performance, cost, and ethics.
\newblock \emph{International Journal of Computer Applications}, 2024.
\newblock URL \url{http://dx.doi.org/10.5120/ijca2024924172}.

\bibitem[Rezaei et~al.(2025)Rezaei, Farnia, and Li]{rezaei2025more}
P.~Rezaei, F.~Farnia, and C.~T. Li.
\newblock Be more diverse than the most diverse: Optimal mixtures of generative
  models via mixture-ucb bandit algorithms.
\newblock In \emph{International Conference on Learning Representations}, 2025.

\bibitem[Schwartz et~al.(2002)Schwartz, Ward, Monterosso, Lyubomirsky, White,
  and Lehman]{schwartz2002maximizing}
B.~Schwartz, A.~Ward, J.~Monterosso, S.~Lyubomirsky, K.~White, and D.~R.
  Lehman.
\newblock Maximizing versus satisficing: happiness is a matter of choice.
\newblock \emph{Journal of personality and social psychology}, 2002.

\bibitem[Tang et~al.(2026)Tang, Hung, Peng, and Yen]{tang2026vdar}
Y.-C. Tang, J.-C. Hung, W.-C. Peng, and A.-Z. Yen.
\newblock Vdar-router: Adaptive llms routing via verbalized query difficulty
  analysis retrieval.
\newblock \emph{arXiv preprint arXiv:2607.18098}, 2026.

\bibitem[Team et~al.(2025)Team, Chen, Ci, Du, Feng, Gai, Guo, Han, He, He,
  et~al.]{team2025kling}
K.~Team, J.~Chen, Y.~Ci, X.~Du, Z.~Feng, K.~Gai, S.~Guo, F.~Han, J.~He, K.~He,
  et~al.
\newblock Kling-omni technical report.
\newblock \emph{arXiv preprint arXiv:2512.16776}, 2025.

\bibitem[Zhang et~al.(2025)Zhang, Feng, and You]{zhang2025routerr}
H.~Zhang, T.~Feng, and J.~You.
\newblock Router-r1: Teaching {LLM}s multi-round routing and aggregation via
  reinforcement learning.
\newblock In \emph{The Thirty-ninth Annual Conference on Neural Information
  Processing Systems}, 2025.
\newblock URL \url{https://openreview.net/forum?id=DWf4vroKWJ}.

\end{thebibliography}

\clearpage
\appendix
\startcontents[appendix]

\part*{Appendices}

\begingroup
\hypersetup{hidelinks}
% Include sections and subsections in the appendix TOC.
\printcontents[appendix]{}{1}[2]{}
\endgroup

\clearpage

\section{Implementation Details}
\label{app:impl}
\paragraph{Encoder and context features.}
Requests are embedded once by the frozen \texttt{Qwen3-VL-Embedding-2B} encoder
(2048-D, prefixed with the task name; I2V requests include the reference
image). The generalized five-fold routers project the context to 32 dimensions
with a PCA fitted on each fold's training set; the personalized routers use the
raw embedding. Generator identity is a family embedding plus an
endpoint-specific residual; candidate features are the normalized context,
generator and user vectors together with their three pairwise products.

\paragraph{Training.}
AdamW with weight decay $0.01$, batch size 64 for pairs and 32 for rubric
rows, temperature $\tau=1$, user dropout $0.4$, up to 50 epochs with early
stopping (patience 10). The general and the personalized ranking stream are
iterated in lockstep, so the number of ranking updates does not depend on the
quality weight. Generalized routers: learning rate selected per fold from
$\{5\times10^{-5},10^{-4},2\times10^{-4}\}$ by inner-validation Top-1 then
NDCG. Personalized routers and the pooled baseline: learning rate
$3\times10^{-4}$, checkpoint with the lowest validation ranking loss (both ranking
streams weighted equally for the personalized routers; the general stream only
for the pooled baseline). Seeds 13, 29 and 47.

\paragraph{Quality weight.}
$\lambda$ was swept over $\{0,10^{-4},10^{-3},0.25\}$ with the ranking step
budget held fixed across $\lambda$. Any positive weight trains the quality head
(test quality AUC $0.79$ for the generalized router versus $0.51$ untrained)
without a measurable change in ranking accuracy; $\lambda=0.25$ gave the best
validation Top-1 and is used for every model. Quality prediction is claimed for
the generalized router only; the personalized routers keep the term for a
uniform objective.

\paragraph{Personalization protocol.}
For each seed, 15\% of prompt groups are held out for validation and 15\% for
test, identically for all nine annotators. The support order of an annotator
is \emph{centroid-diverse}: starting from the questionnaire embedding, the next
prompt group is the one whose centroid embedding is least similar to the
centroid of the groups already selected. A cap $c$ keeps the first $c$ groups;
the matched pooled model at cap $c$ takes $c$ groups in total in balanced
round-robin order over annotators. Reported curves use caps
$\{1,2,5,10,20,40,50\}$ and the full data. Same-user kNN uses $k=20$.

\paragraph{Agentic baselines.}
Octo keeps a capability memory of per-generator pass rates over prompt-derived
capability buckets built from training-fold verifier outcomes, routes to the
cheapest generator whose expected pass rate exceeds a threshold (threshold and
minimum bucket count selected on validation), and escalates to the
next-cheapest generator for up to four attempts when the verifier rejects the
output. The verifier rates visual quality, prompt alignment and physical
consistency on a 1--5 scale and accepts a video when the mean is at least
$4.0$ and every dimension at least $3.0$. FrugalGPT learns, per budget, a
cascade order of up to four generators and per-stage acceptance thresholds on
the training split, selects the configuration on validation, and replays it
frozen at test time under a per-pool cumulative cost cap; the last affordable
stage always returns its output. Both use the verifier's mean rubric score as
the answer scorer; generation cost accumulates over attempts and verifier API
or compute cost is reported separately. kNN was tuned over
$k\in\{1,10,20,30,50,100\}$.

\paragraph{Costs.}
Generation cost is the generator's catalog price per video at the time of
collection, known before generation; prices are frozen for all experiments.

\section{Additional Dataset Analysis}
\label{app:analysis}
This appendix collects the dataset analyses that Section~\ref{sec:analysis} summarizes in
one or two sentences. Every estimator, seed and input is the same as for the
corresponding figure in the main text.

\paragraph{Which generators win, and how often.}
Table~\ref{tab:annotation_generator} lists the five most frequent pool-level winners.
The distribution is flat: the leader takes 10.1\% of the pools, nine of the 30 generators
each take more than 5\%, and 28 win at least one pool, which is why a single default
generator is a weak policy and routing has something to exploit.

\begin{table}[tb]
\caption{Distribution of pool-level winning generators by consensus utility. Each of the 393 evaluable pools contributes one vote, with ties divided fractionally among co-winners. The five most frequent winners are shown; nine of the 30 generators each win more than 5\% of the pools and 28 win at least one.}
\label{tab:annotation_generator}
\begin{center}
\begin{small}
\begin{tabular}{lcc}
\toprule
\bf Generator & \bf Fractional pool wins & \bf Win rate \\
\midrule
       ltx-2.3 & 39.7 & 10.10\% \\
       veo-3.1-fast & 34.8 & 8.84\% \\
       veo-3.1-lite-generate-001 & 31.2 & 7.95\% \\
       seedance-2.0-mini & 28.8 & 7.32\% \\
       veo-3.1-lite & 28.2 & 7.18\% \\
\bottomrule
\end{tabular}
\end{small}
\end{center}
\vskip -0.5cm
\end{table}

\begin{figure}[tb]
  \centering
  \includegraphics[width=0.70\textwidth]{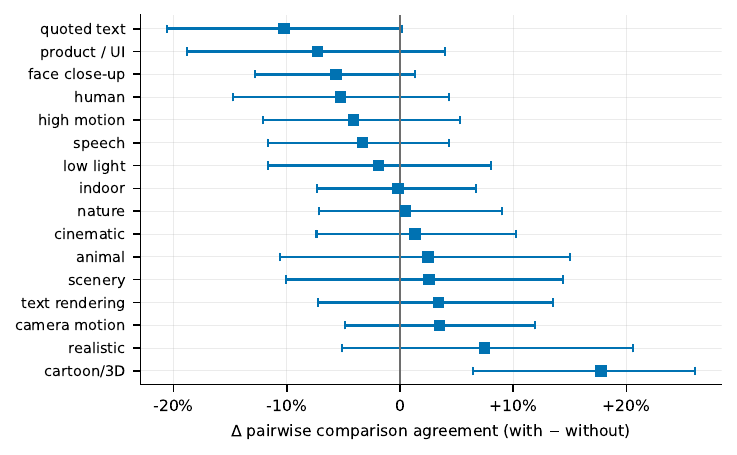}
  \caption{\textbf{Prompt attributes barely move agreement.} Among clean pairs (T2V and
  I2V pooled), the change in preference agreement when a keyword-detected prompt
  attribute is present versus absent, with pool-level bootstrap 95\% intervals, since
  comparisons within a pool are correlated. Only cartoon or 3D prompts have an interval
  that excludes zero.}
  \label{fig:agreement_attributes}
\end{figure}

\paragraph{Prompt attributes do not explain who agrees with whom.}
Figure~\ref{fig:agreement_attributes} gives the full version of the attribute effect
quoted in Section~\ref{sec:analysis}: for each of the 16 keyword-detected attributes, the
change in preference agreement among clean pairs when the attribute is present. Only
cartoon or 3D prompts have an interval that excludes zero, so prompt content explains
little of the disagreement that remains once both videos are acceptable.

\begin{figure}[tb]
  \centering
  \includegraphics[width=0.53\textwidth]{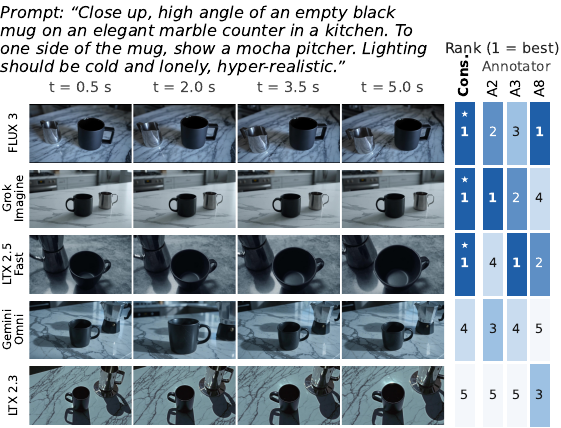}
  \caption{\textbf{Cyclic favorites.} A T2V pool in which no annotator rated any video
  Not acceptable, laid out as in Figure~\ref{fig:disagreement_consensus}, where
  $\star$ marks a rank shared by more than one generator. A2 prefers
  Grok Imagine, A3 LTX~2.5~Fast, and A8 FLUX~3; the majority relation is cyclic
  (Grok $\succ$ FLUX $\succ$ LTX~2.5~Fast $\succ$ Grok), and the consensus ties all
  three for first.}
  \label{fig:disagreement_cycle}
\end{figure}

\paragraph{A pool with no possible consensus winner.}
Figure~\ref{fig:disagreement_cycle} is the second of the two qualitative examples in
Section~\ref{sec:analysis}. Three annotators accept every video and each names a
different favorite, and their majority relation is cyclic, so no aggregation rule can
produce a winner that more than one of them endorses. Cycles of this kind occur in 7\%
of all pools.

\section{Additional Routing Results}
\label{app:metrics}
This appendix carries the unabridged version of the routing figure and the two metrics the
main text leaves out. Colors and markers follow the main text, and every panel uses the
same full-candidate ground truth.

\begin{figure}[h]
  \centering
  \includegraphics[width=\textwidth]{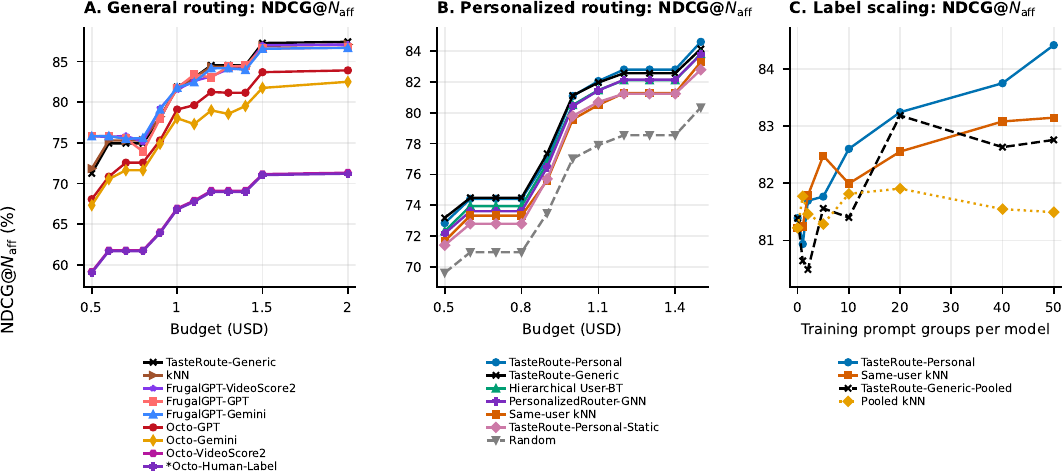}
  \caption{\textbf{NDCG across the three routing experiments, all methods}, laid out as
  Figure~\ref{fig:routing-ndcg}.}
  \label{fig:routing-ndcg-full}
\end{figure}

\paragraph{All methods on NDCG.}
Figure~\ref{fig:routing-ndcg-full} repeats Figure~\ref{fig:routing-ndcg} with every series:
the three verifier variants of each agentic family, the held-out-label oracle, and the
random baseline. The variants within a family differ by less than a point of NDCG, which is
why the main text shows one of each. Octo-Human-Label reads the held-out labels, so it is a
diagnostic ceiling rather than a method; on NDCG it sits at the bottom because a cascade
never orders the candidates it did not generate.

\begin{figure}[h]
  \centering
  \includegraphics[width=\textwidth]{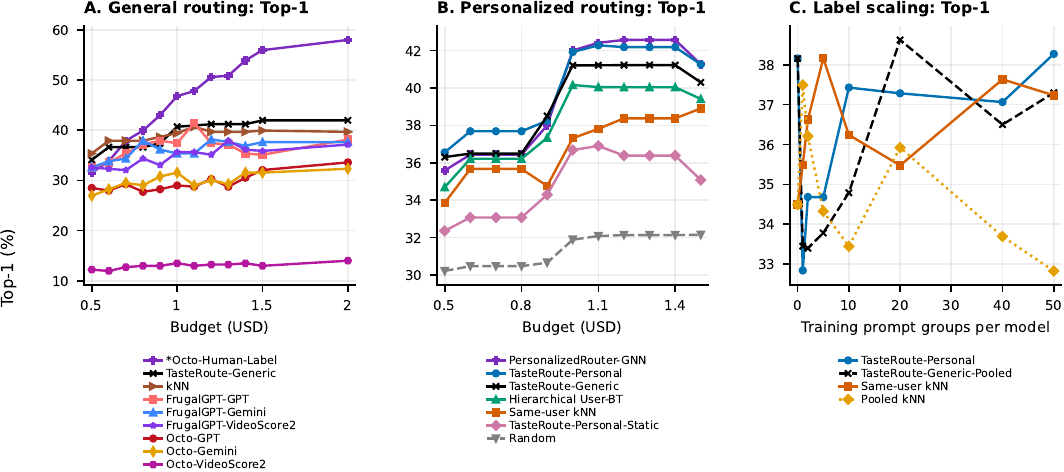}
  \caption{\textbf{Top-1 across the three routing experiments}, laid out as
  Figure~\ref{fig:routing-ndcg}. (A)~General routing by budget. (B)~Personalized
  routing by budget. (C)~Matched-budget label scaling.}
  \label{fig:routing-top1}
\end{figure}

\begin{figure}[h]
  \centering
  \includegraphics[width=\textwidth]{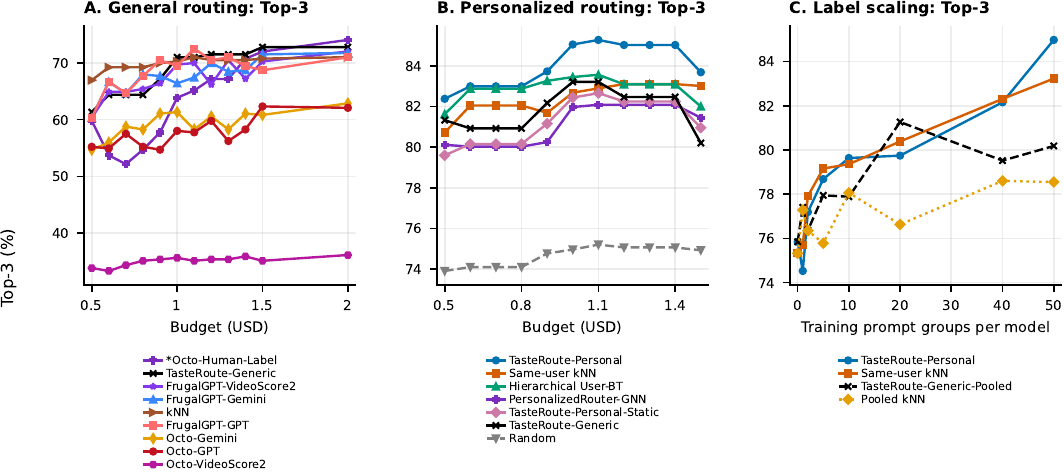}
  \caption{\textbf{Top-3 across the three routing experiments}, laid out as
  Figure~\ref{fig:routing-ndcg}. (A)~General routing by budget. (B)~Personalized
  routing by budget. (C)~Matched-budget label scaling.}
  \label{fig:routing-top3}
\end{figure}

\paragraph{Top-1 and Top-3.}
Top-1 and Top-3 over the same runs are below; the panels, budgets and baselines are
identical, and every conclusion in Section~\ref{sec:res} holds on all three metrics. Figure~\ref{fig:routing-top1} gives
Top-1 and Figure~\ref{fig:routing-top3} gives Top-3. Top-1 is the noisiest of the three,
since a single held-out pool contributes one binary outcome; Top-3 is the most saturated,
with every learned method above 80\% at most budgets, which is why the main text reports
NDCG.

\section{Showcase of Labeling UI}
\label{app:labeling_ui}

\begin{figure}[h]
  \centering
  \includegraphics[width=0.78\textwidth]{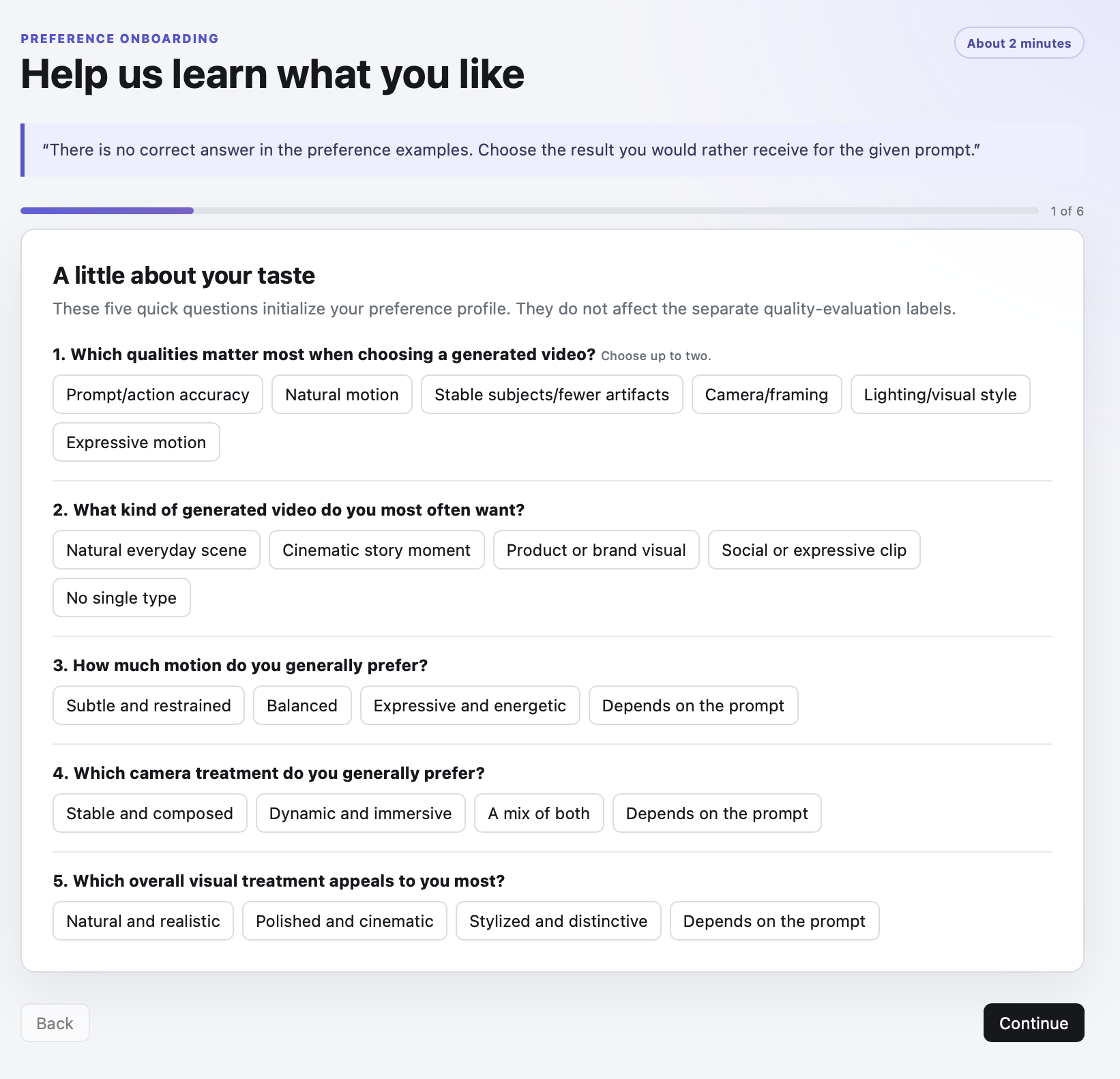}
  \caption{Onboarding step 1: the five-item preference questionnaire. Responses initialize the annotator's preference profile and do not affect the separate quality-evaluation labels.}
  \label{fig:ui_questionnaire}
\end{figure}

\begin{figure}[h]
  \centering
  \includegraphics[width=0.78\textwidth]{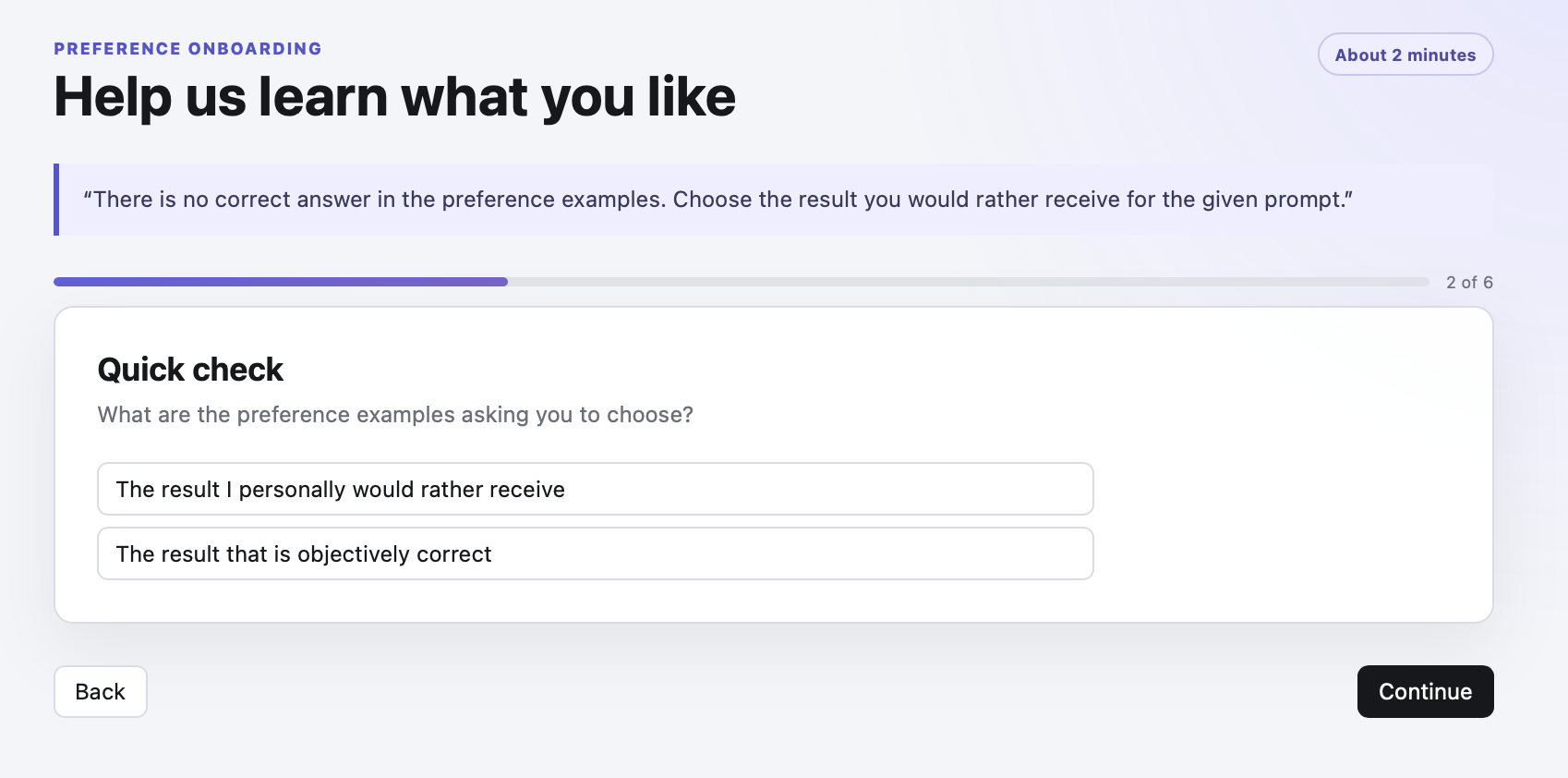}
  \caption{Onboarding step 2: comprehension check confirming that annotators understand the preference examples ask for the result they personally would rather receive, not an objectively correct one.}
  \label{fig:ui_check}
\end{figure}

\begin{figure}[h]
  \centering
  \includegraphics[width=0.78\textwidth]{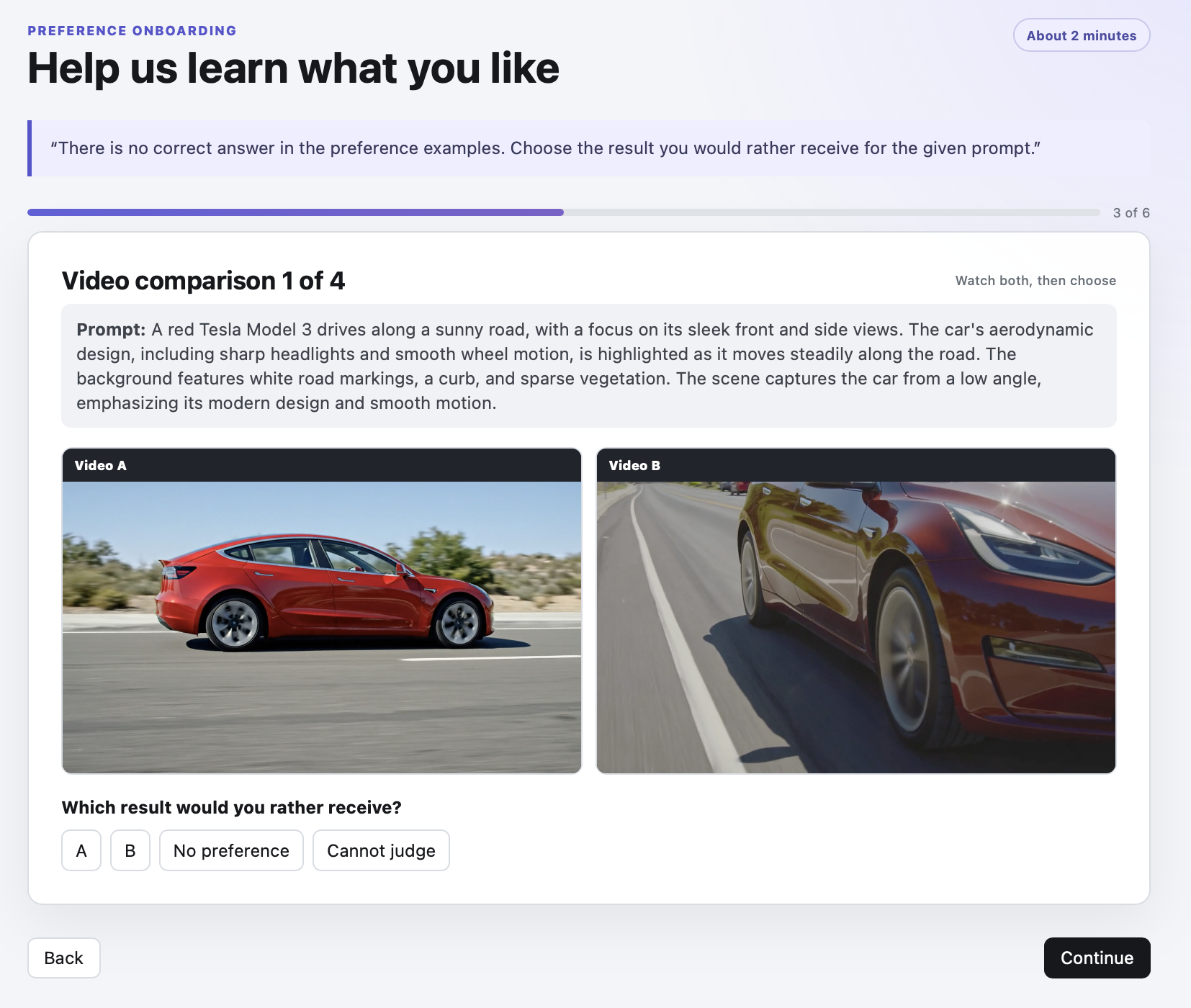}
  \caption{Onboarding steps 3--6: a reference-video comparison. Annotators watch both videos and choose which result they would rather receive.}
  \label{fig:ui_comparison}
\end{figure}

\begin{figure}[h]
  \centering
  \begin{minipage}{0.78\textwidth}
    \centering
    \includegraphics[width=\linewidth]{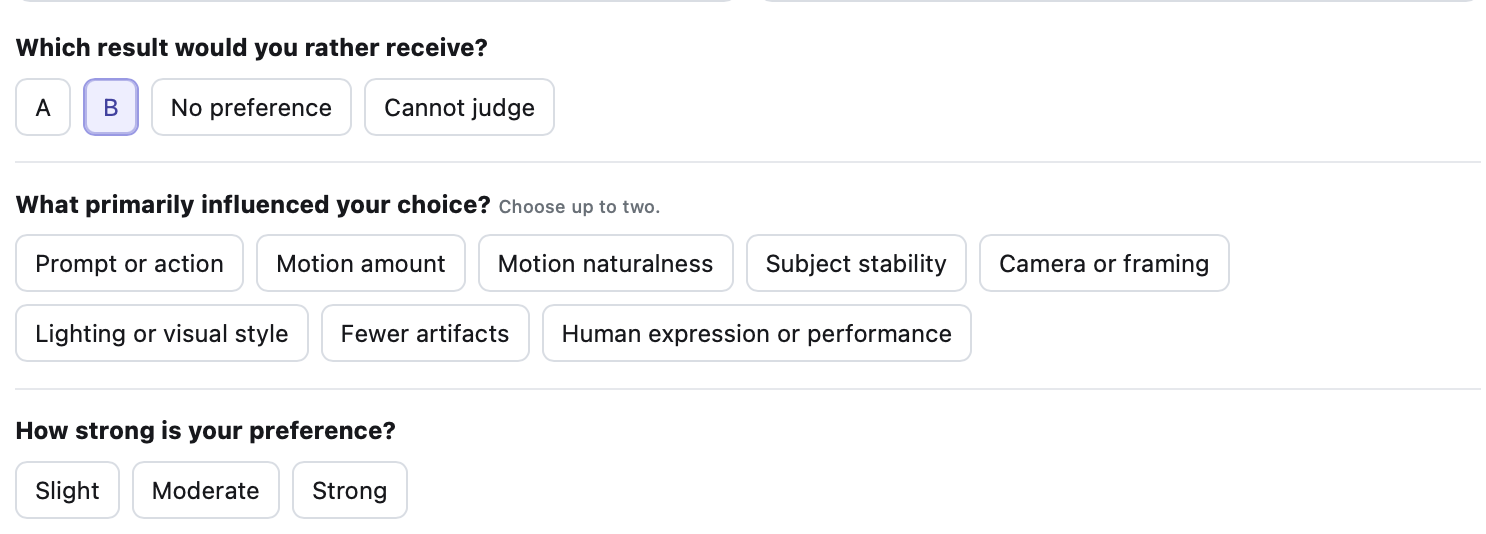}\\
    \small (a) Directional choice (A or B): influencing factors and preference strength.\\[6pt]
    \includegraphics[width=\linewidth]{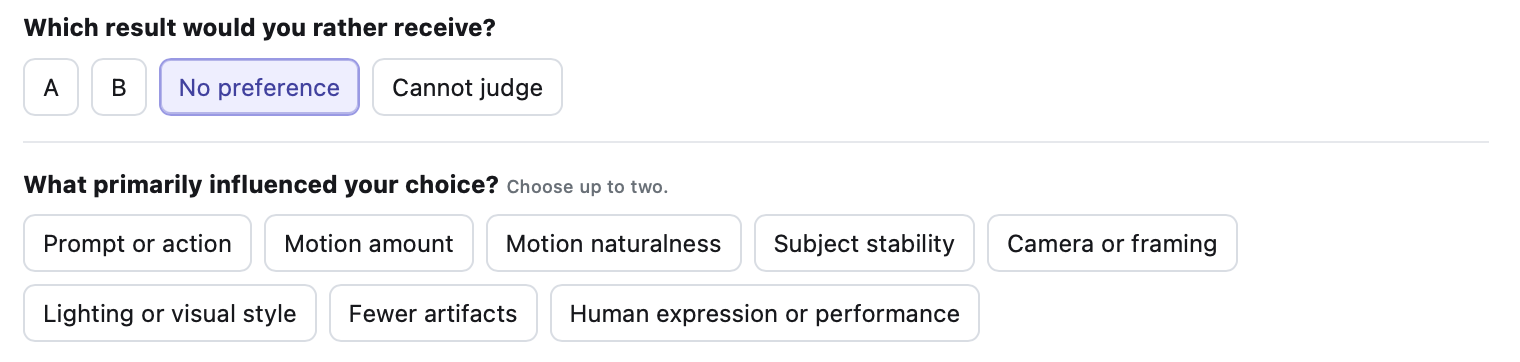}\\
    \small (b) No preference: influencing factors only.\\[6pt]
    \includegraphics[width=\linewidth]{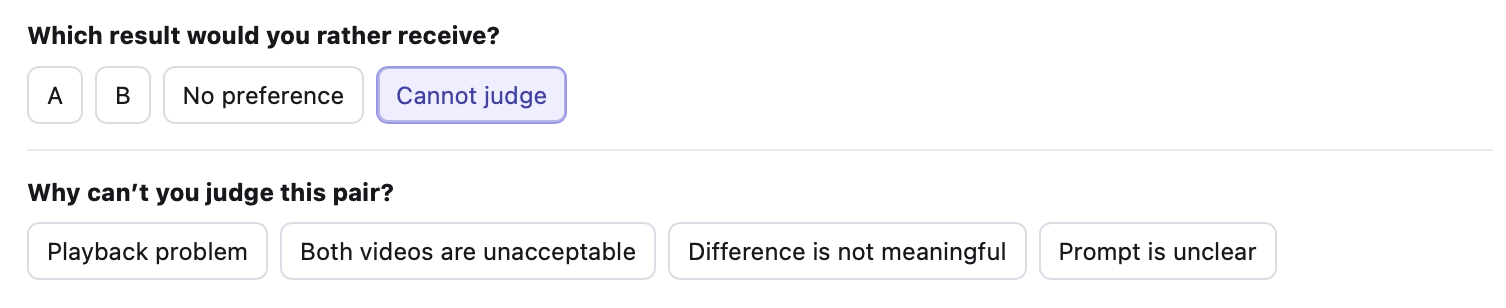}\\
    \small (c) Cannot judge: reason for abstaining.
  \end{minipage}
  \caption{Follow-up questions shown after each choice in a reference-video comparison.}
  \label{fig:ui_followups}
\end{figure}

\begin{figure}[h]
  \centering
  \includegraphics[width=\textwidth]{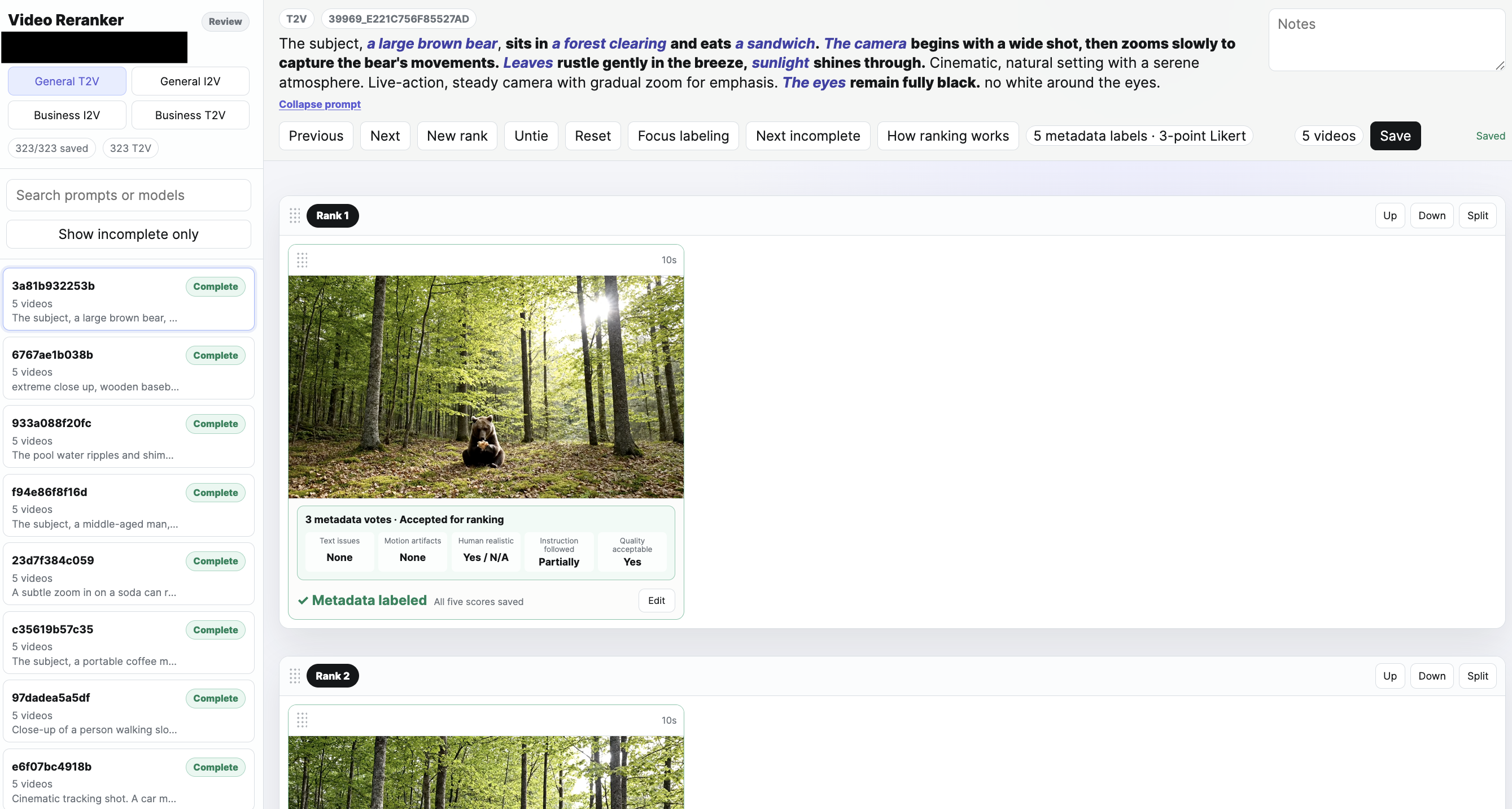}
  \caption{The ranking interface. Each video card shows its quality ratings; videos are ordered into rank groups, with ties allowed. Annotator-identifying information is redacted.}
  \label{fig:ui_ranking}
\end{figure}

We show the web interface used to collect TasteRoute-3K (Section~\ref{sec:data}). Each annotator first completes a six-step preference onboarding that initializes their preference profile (Figures~\ref{fig:ui_questionnaire}--\ref{fig:ui_followups}), and then labels prompts in a ranking interface that collects both per-video quality ratings and listwise preference rankings (Figure~\ref{fig:ui_ranking}).

\paragraph{Preference onboarding.}
The onboarding opens with a five-item questionnaire on self-reported preferences: the qualities that matter most (up to two), the preferred video type, motion amount, camera treatment, and overall visual treatment (Figure~\ref{fig:ui_questionnaire}). Throughout onboarding, a banner reminds annotators that there is no correct answer and that they should choose the result they would rather receive. To make this framing explicit, annotators then answer a quick comprehension check asking what the preference examples ask them to choose (Figure~\ref{fig:ui_check}).

\paragraph{Reference-video comparisons.}
Because self-reported preferences may not reflect actual choices, the remaining onboarding steps present four reference-video comparisons. Annotators watch two generations for the same prompt and choose A, B, no preference, or cannot judge (Figure~\ref{fig:ui_comparison}). The follow-up questions depend on the choice (Figure~\ref{fig:ui_followups}): selecting A or B asks for up to two factors that influenced the choice and the preference strength (slight, moderate, strong); selecting no preference asks only for the influencing factors; and selecting cannot judge asks for a reason (playback problem, both videos unacceptable, difference not meaningful, or unclear prompt).

\paragraph{Ranking interface.}
After onboarding, annotators label prompts in the ranking interface (Figure~\ref{fig:ui_ranking}). The left panel lists prompts by task (general or business, T2V or I2V) with their completion status, and the top of the main panel shows the conditioning prompt with its key entities highlighted. For each video, annotators first provide the five quality fields (text issues, motion artifacts, human realism, instruction following, and overall acceptability) on a 3-point scale. They then arrange the candidate videos into rank groups by dragging or using the up/down controls; placing several videos in one group expresses a tie, and a group can be split to break it.

\section{Video Generators and Cost}
\label{app:models}

Table~\ref{tab:model_list} lists the 30 video generators in the TasteRoute-3K candidate pool (Section~\ref{sec:data}), ordered by price. Prices are the catalog rates at the time of collection, in USD per second of 720p video without audio, and are frozen for all experiments (Appendix~\ref{app:impl}). Some models are listed under two endpoint IDs, and each ID is treated as a separate candidate.

\begin{table}[h]
\centering
\small
\caption{Video generators in the candidate pool. \emph{Tasks} gives the settings in which each generator appears in TasteRoute-3K. Prices are in USD per second of 720p video without audio; Veo and \texttt{gemini-omni-flash} prices follow Google's API catalogs, and all others follow fal.ai.
$^\dagger$No 720p tier; billed at the 1080p rate.
$^\ddagger$Flat \$0.15 per video, converted at 6\,s.
$^\S$Base price covers 5\,s plus a per-second rate; shown as the 6\,s equivalent.}
\label{tab:model_list}
\begin{tabular}{llcr}
\toprule
Model & Endpoint ID & Tasks & \$/s \\
\midrule
LongCat Video Distilled & \texttt{longcat-video-distilled-720p} & T2V & 0.010 \\
Wan 2.2 5B & \texttt{wan-v2.2-5b} & T2V, I2V & 0.025$^\ddagger$ \\
Veo 3.1 Lite & \texttt{veo-3.1-lite} & T2V, I2V & 0.030 \\
Veo 3.1 Lite & \texttt{veo-3.1-lite-generate-001} & T2V & 0.030 \\
LTX Video 13B Distilled & \texttt{ltx-video-13b-distilled} & T2V, I2V & 0.040 \\
Pika 2.2 & \texttt{pika-v2.2} & T2V, I2V & 0.040 \\
Kling 2.5 Turbo Standard & \texttt{kling-v2.5-turbo-standard} & I2V & 0.042$^\S$ \\
PixVerse V6 & \texttt{pixverse-v6} & T2V & 0.045 \\
PixVerse C1 & \texttt{pixverse-c1} & I2V & 0.050 \\
LTX 2.3 Fast & \texttt{ltx-2.3-fast} & T2V, I2V & 0.060$^\dagger$ \\
Luma Ray 3.2 & \texttt{luma-ray-v3.2} & T2V, I2V & 0.060 \\
Grok Imagine Video & \texttt{grok-imagine-video} & T2V & 0.070 \\
Kling 2.5 Turbo Pro & \texttt{kling-v2.5-turbo-pro} & T2V & 0.070$^\S$ \\
Vidu Q3 Turbo & \texttt{vidu-q3-image-to-video-turbo} & I2V & 0.077 \\
Vidu Q3 Turbo & \texttt{vidu-q3-text-to-video-turbo} & T2V & 0.077 \\
LTX 2.3 Pro & \texttt{ltx-2.3} & T2V, I2V & 0.080$^\dagger$ \\
Veo 3.1 Fast & \texttt{veo-3.1-fast} & T2V, I2V & 0.080 \\
Veo 3.1 Fast & \texttt{veo-3.1-fast-generate-001} & T2V & 0.080 \\
Wan 2.2 A14B & \texttt{wan-v2.2-a14b} & T2V, I2V & 0.080 \\
LTX 2.5 Fast & \texttt{ltx-2.5-fast} & T2V & 0.090 \\
Gemini Omni Flash & \texttt{gemini-omni-flash} & T2V, I2V & 0.100 \\
Gemini Omni Flash 1.1 & \texttt{gemini-omni} & T2V & 0.100 \\
Grok Imagine Video 1.5 & \texttt{grok-imagine} & T2V & 0.140 \\
Grok Imagine Video 1.5 & \texttt{grok-imagine-video-v1.5} & I2V & 0.140 \\
Happy Horse 1.1 & \texttt{happy-horse-v1.1} & T2V, I2V & 0.140 \\
Seedance 2.0 Mini & \texttt{seedance-2.0-mini} & T2V, I2V & 0.1547 \\
FLUX 3 & \texttt{flux-3} & T2V & 0.170 \\
Veo 3.1 & \texttt{veo-3.1} & T2V & 0.200 \\
Veo 3.1 & \texttt{veo-3.1-generate-001} & T2V, I2V & 0.200 \\
Seedance 2.0 Fast & \texttt{seedance-2.0-fast} & T2V, I2V & 0.2419 \\
\bottomrule
\end{tabular}
\end{table}

\end{document}